\documentclass{article}

\usepackage[final]{neurips_data_2024}

\usepackage[utf8]{inputenc} 
\usepackage[T1]{fontenc}    
\usepackage{hyperref}       
\usepackage{url}            
\usepackage{booktabs}       
\usepackage{amsfonts}       
\usepackage{nicefrac}       
\usepackage{microtype}      
\usepackage{xcolor}         
\usepackage{wrapfig}
\usepackage{amsmath}
\usepackage{graphicx}
\usepackage{multirow}
\usepackage{bbding}
\usepackage{floatrow}
\newfloatcommand{capbtabbox}{table}[][\FBwidth]

\definecolor{redrock}{HTML}{c1282e}
\definecolor{greengravel}{HTML}{029e73}
\definecolor{yellowsand}{HTML}{d5b60a}
\definecolor{bluemud}{HTML}{006ca8}

\usepackage[breakable, most]{tcolorbox}
\AtBeginEnvironment{tcolorbox}{\small}
\definecolor{myframecolor}{HTML}{7f7f7f}
\definecolor{mycolback}{HTML}{e2e2e2}

\usepackage{textcomp}
\usepackage{fancyvrb}

\usepackage{listings}
\title{SeafloorAI: A Large-scale Vision-Language Dataset \\for Seafloor Geological Survey}

\author{%
  Kien X. Nguyen$^1$, Fengchun Qiao$^1$, Arthur Trembanis$^2$, Xi Peng$^1$ \\
  $^1$\href{https://deep-real.github.io/}{Deep-REAL Lab}, Department of Computer and Information Sciences, University of Delaware \\
  $^2$School of Marine Science and Policy, University of Delaware \\
  \texttt{\{kxnguyen,fengchun,art,xipeng\}@udel.edu}
}

\begin{document}

\maketitle

\begin{abstract}\label{sec:abs}
  A major obstacle to the advancements of machine learning models in marine science, particularly in sonar imagery analysis, is the scarcity of AI-ready datasets.
  While there have been efforts to make AI-ready sonar image dataset publicly available, they suffer from limitations in terms of environment setting and scale.
  To bridge this gap, we introduce \texttt{SeafloorAI}, the first extensive AI-ready datasets for seafloor mapping across 5 geological layers that is curated in collaboration with marine scientists. We further extend the dataset to \texttt{SeafloorGenAI} by incorporating the language component in order to facilitate the development of both \textit{vision}- and \textit{language}-capable machine learning models for sonar imagery.
  The dataset consists of 62 geo-distributed data surveys spanning 17,300 square kilometers, with 696K sonar images, 827K annotated segmentation masks, 696K detailed language descriptions and approximately 7M question-answer pairs. 
  By making our data processing source code publicly available, we aim to engage the marine science community to enrich the data pool and inspire the machine learning community to develop more robust models. 
  This collaborative approach will enhance the capabilities and applications of our datasets within both fields. 
  Our code repository are available~\footnote{\url{https://github.com/deep-real/SeafloorAI}} under the CC-BY-4.0 license.
\end{abstract}

\section{Introduction}

\begin{figure}[!htb]
    \centering
    \includegraphics[width=\linewidth]{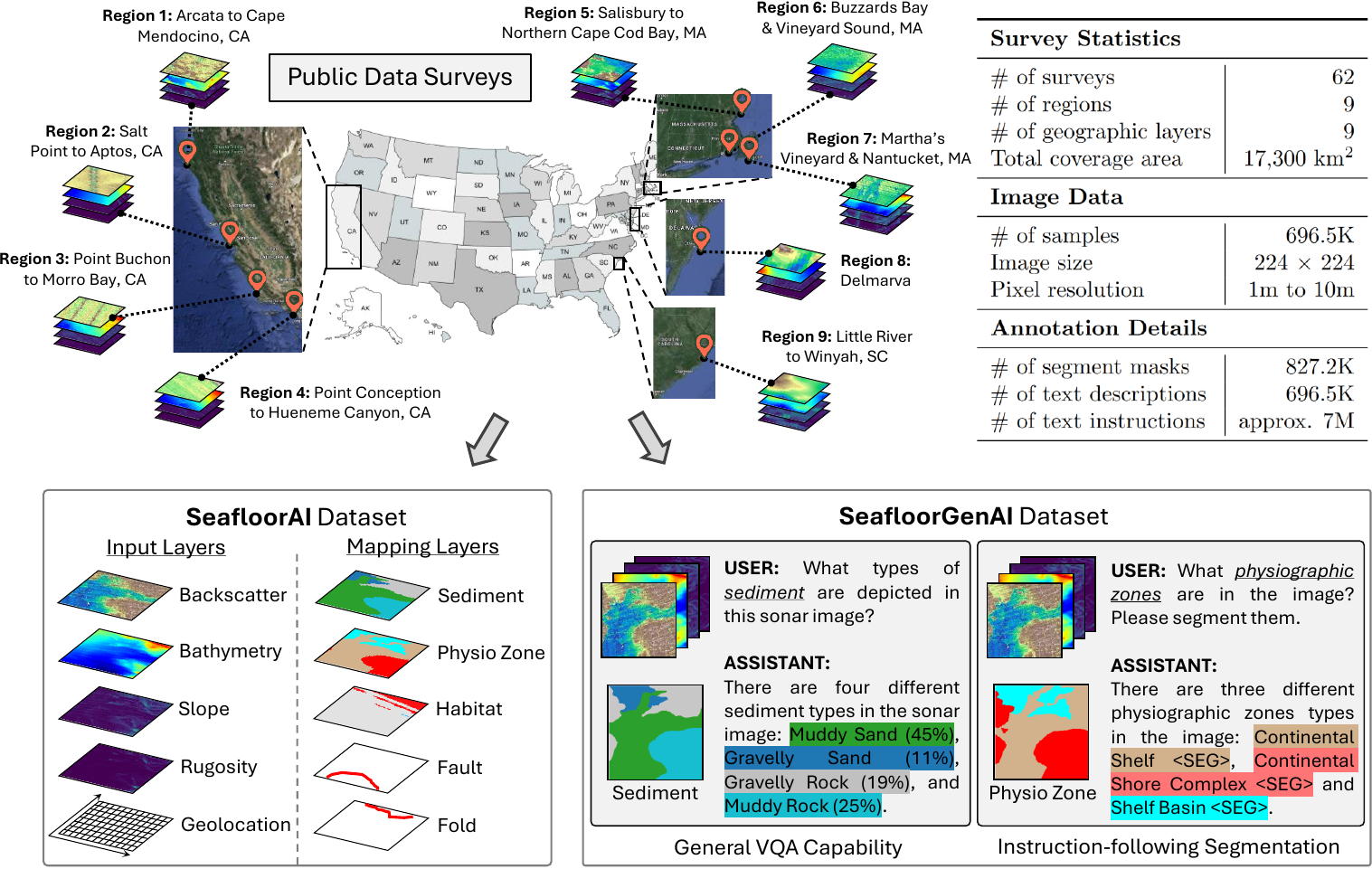}
    \caption{Overview of the spatially distributed seafloor mapping datasets. The table highlights key dataset statistics. We incorporate 62 public data surveys published by USGS and NOAA from 9 major regions to construct \texttt{SeafloorAI} and \texttt{SeafloorGenAI} datasets. Our dataset contains 9 geological layers, 4 of which are raw signals, \textit{i.e.}, Backscatter, Bathymetry, Slope and Rugosity, and 5 annotated by human experts, \textit{i.e.} Sediment, Physiographic Zone, Habitat, Fault and Fold. \texttt{SeafloorAI} serves as a dataset for standard computer vision tasks, \textit{i.e.} semantic segmentation, whereas \texttt{SeafloorGenAI} constitutes a dataset for generative vision-language tasks, \textit{i.e.}, general visual question answering and instruction-following mapping. \texttt{<SEG>} denotes the segmentation mask output by the model.}
    \label{fig:title}
\end{figure}

Seafloor mapping stands at the forefront of marine science, utilizing cutting-edge technologies like multibeam echosounders and side-scan sonar to unveil the hidden complexities of the ocean floor~\cite{Trembanis2013Detailed,Trembanis2020-cl}. 
Beyond scientific research, seafloor mapping is instrumental in identifying potential resources, assessing environmental impacts, and supporting sustainable ocean management practices in the context of the blue economy~\cite{Mayer2018-cx}. 
However, the current analysis techniques in seafloor mapping are predominantly labor-intensive and reliant on manual interpretation by marine scientists, necessitating hundreds of hours spent meticulously examining data surveys to analyze seabed imagery~\cite{Trembanis2019-gs}.
This hands-on approach is not only time-consuming but also susceptible to user \textit{subjectivity} and the limitations of individual expertise, thus introducing potential \textit{inconsistencies} in analysis~\cite{Raineault2012-py}.

The integration of machine learning (ML) holds the promise of enhancing efficiency and reliability in seafloor mapping by automating the segmentation and classification tasks~\cite{arosio2023fully,qin2021ieee,lundine2023deep}.
However, the lack of public AI-ready datasets poses a significant challenge in leveraging the full potential of AI technologies for this purpose.
While there have been efforts to make AI-ready sonar image datasets publicly available, they suffer from limitations in terms of environment setting and scale.
For example, the dataset in~\cite{Singh2021TheMD} was captured in a water tank, which does not accurately represent the ocean's complex conditions. 
Additionally, other work have only produced small-scale datasets with limited area coverage~\cite{luo2019ieee,Sethuraman2024MachineLF}, not accounting for the generalizability of the ML models in a spatially distributed setting.
On the other end, abundant public hydrographic surveys conducted by the U.S. Geological Survey (USGS) and the National Oceanographic and Atmospheric Administration (NOAA)\footnote{Provides public domain data license.}~\cite{Golden2013-kn,Pendleton2019-ci, Pendleton2015Nahant,Pendleton2013CapeCod, Pendleton2018Nantucket,Ackerman2015-bk,Baldwin2004SC} have yet to be extensively utilized by the ML community.

To bridge this gap, we introduce \texttt{SeafloorAI}, the first extensive AI-ready sonar imagery dataset for seafloor mapping.
We compiled 62 public hydrographic surveys to construct a large-scale, geo-distributed and multi-purpose dataset, with the effort to map various geological layers.
Furthermore, inconsistencies in the nomenclature of geological attributes across data surveys pose a challenge on the unification and development of an extensive dataset.
In collaboration with marine scientists, we have developed a framework that standardizes such nomenclature by adopting the Barnhardt classification~\cite{Barnhardt98} and the Coastal and Marine Ecological Classification Standard (CMECS)~\cite{noaaCoastalMarine}. It guarantees uniformity throughout the dataset, enabling the evaluation of robust ML models in a spatially distributed setting.
The data pool currently consists of 696K sonar images, 827K segmentation masks for 5 geological layers: Sediment, Physiographic Zone, Habitat, Fault, and Fold.

\begin{table}[t!]
\centering

\resizebox{\linewidth}{!}{
    \begin{tabular}{l|c|c|r|rrrrr|r}
    \toprule
    \multirow{2}{*}{\textbf{Location}} & \multirow{2}{*}{\shortstack[c]{\textbf{Region}\\ \textbf{Index}}}  & 
    \multirow{2}{*}{\shortstack[c]{\textbf{Image}\\ \textbf{Resolution}}} &\multirow{2}{*}{\shortstack[c]{\textbf{Input}\\ \textbf{Layers}}} & \multicolumn{5}{c|}{\textbf{Mapping Layers}} & \multirow{2}{*}{\shortstack[c]{\textbf{Area}\\(km$^2$)}} \\
    & & & & \multicolumn{1}{c}{Sediment} & \multicolumn{1}{c}{Physio Zone} & \multicolumn{1}{c}{Habitat} & \multicolumn{1}{c}{Fault} & \multicolumn{1}{c|}{Fold} &  \\
    \midrule
        \multirow{4}{*}{California} & Region 1 & 2m/pixel & 25,817 & 25,672& 25,823& & &  & 672\\
        & Region 2 & 2m/pixel & 123,774& & &  123,480& 123,774&123,774 & 3,148 \\
        & Region 3 & 2m/pixel & 21,270& 20,861& 21,253& & & & 564 \\
        & Region 4 & 2m/pixel & 42,771& & &  25,579& 42,771& 42,771 & 1,419 \\
    \midrule
        \multirow{3}{*}{Massachussetts} & Region 5 & 10m/pixel & 15,827& 4,647& 3,387&  & & & 5,496 \\
        & Region 6 & 1m/pixel & 122,441&  122,236& 118,175& & & & 228\\
        & Region 7 & 1m/pixel & 1,593&  1,507& 1,510& & & & 454 \\
    \midrule
        Delmarva & Region 8 & 2m/pixel & 329,881& & & & & & 4,525 \\
    \midrule
        South Carolina & Region 9 & 4m/pixel & 13,141& & & & & & 808 \\
    \midrule
        \multicolumn{3}{r|}{\textbf{Total}} & 696,515 &  174,923& 170,148& 149,059& 166,545& 166,545 & 17,314\\
    \bottomrule
    \end{tabular}
}
    \caption{Summary of the seafloor mapping data available for each region. The input layers for sonar images include Backscatter, Bathymetry, Slope and Rugosity. Due to different mapping objectives of the original data surveys, the availability of segmentation masks is not uniform across mapping layers. Regions with unlabeled data can be utilized to pre-train the model via self-supervised learning~\cite{gui2024surveyselfsupervisedlearningalgorithms}.}
    \label{tab:surveys}
    \vspace{-5pt}
\end{table}

Finally, we incorporate the language component into our dataset for the development of generative vision-language models (VLMs) in marine science research. 
VLMs facilitate seamless interactions through textual queries and provide clear, understandable explanations throughout the analysis process~\cite{li2024llava,li2024deal}. 
In addition, the ability to automate a report of the survey's findings, such as sediment composition, habitats, \textit{etc.}, would reduce the time and effort required for manual preparation. 
To this end, we present a data curation pipeline that leverages both domain knowledge from marine scientists and language generation capability of \texttt{GPT-4}~\cite{openai2024gpt4}. 
Specifically, we employ in-context learning~\cite{brown2020language} to generate analysis-driven question-answer pairs for each image, resulting in 7M samples and 696K language descriptions. We name the vision-language dataset \texttt{SeafloorGenAI}.

Our contributions are summarized as follows:\vspace{-3pt}
\begin{enumerate}
    \item We compile 62 public hydrographic data surveys from USGS and NOAA into a large, geo-distributed, multi-purpose and multi-modal dataset for seafloor mapping research.
    \item We provide a standardization of naming convention across these surveys, under the \textit{rigorous supervision of marine scientists}, to unify an extensive AI-ready dataset.
    \item We present a data curation pipeline that produces detailed descriptions and question-answer pairs for the development of large generative vision-language models in marine science.
    \item Our geo-distributed dataset contains 696K sonar images, 827K segmentation masks, 696K language descriptions and 7M question-answer pairs, covering a total area of 17,300 square kilometers.
    \item We open-source our data processing code so that marine scientists could efficiently contribute their data surveys to expand the data pool.
\end{enumerate}

\section{Related Work}
\vspace{-5pt}
\textbf{Underwater Imagery Datasets.}
Over the years, researchers at USGS and NOAA have carried out frequent hydrographic surveys~\cite{Golden2013-kn,Pendleton2019-ci, Pendleton2015Nahant,Pendleton2013CapeCod, Pendleton2018Nantucket,Ackerman2015-bk, Baldwin2004SC} to collect and provide accurate and reliable information about the physical features of the water bodies and the seafloor. 
They are instrumental in creating accurate nautical charts to identify underwater hazards, aiding in the planning of marine infrastructure, and providing essential data for scientific research and environmental conservation. 
Furthermore, the data supports various economic activities, such as fishing, aquaculture, and energy production, by enabling sustainable and efficient operations. 
In recent years, substantial efforts have been made to create public AI-ready underwater datasets, including forward-looking sonar (FLS), side-scan sonar (SSS), and RGB imagery. These datasets are utilized to develop machine learning models tailored for domain applications, focusing on classification or detection of geological features~\cite{arosio2023fully,chen2016deep,berthold2017,luo2019ieee,qin2021ieee,lundine2023deep} and man-made objects~\cite{ye2018ieee,wang2019underwater,huo2020ieee,xu2020underwater,li2021zero,nayak2021machine,cheng2022multi,williams2019transfer,williams2016underwater,mckay2017s,zhu2018underwater,warakagoda2018transfer}.
Singh and Valdenegro-Toro~\cite{Singh2021TheMD} were pioneers with their FLS image dataset aimed at object detection, but their use of a controlled water tank setting may not fully reflect the complex oceanic conditions, limiting the generalizability of their results. 
Xie et al.~\cite{Xie2022ADW} addressed this by extending object detection to data collected in natural water bodies, enhancing its real-world applicability. 
Sethuraman et al.~\cite{Sethuraman2024MachineLF} developed an SSS dataset for shipwreck detection, though its small sample size could limit model robustness.
Others have also explored RGB underwater imagery for trash detection~\cite{Walia2023OptimizedCD} and semantic segmentation~\cite{Islam2020Semantic}.

Our research focuses on transforming the USGS and NOAA hydrographic surveys into a comprehensive, multi-scale, multi-purpose and multi-modal SSS imagery dataset. 
This initiative aims to propel advancements in both marine science and machine learning research, creating a bridge between extensive marine data resources and innovative computational techniques.

\vspace{5pt}
\textbf{Why side-scan sonar?} Compared to FLS and RGB imagery, SSS offers distinct advantages for underwater imagery analysis. 
Side-scan sonar provides a wider coverage area, and creates high-resolution images that clearly delineate the seabed texture, which is essential for geological surveys, shipwreck location, and habitat mapping.
Unlike FLS, which is primarily used for obstacle avoidance, SSS offers a broad, fan-shaped beam that scans the ocean floor to either side of the towfish or autonomous underwater vehicle, capturing detailed images of the seafloor texture. 
Moreover, SSS is less affected by water turbidity compared to RGB cameras, which struggle with visibility in murky waters and suffer from significant color loss at depth due to light absorption.
This allows SSS to produce consistent and reliable imagery under a variety of underwater conditions, where optical methods would fail.
Still, SSS is only a 2D representation of the seabed. 
We also incorporate 3D information such as water depth to describe the underwater topography.
This allows for a broad scope of underwater imagery analysis, providing robust data suitable for in-depth assessments.

\vspace{5pt}
\textbf{Comparison with Existing Datasets.}
Our dataset is a comprehensive and expansive dataset that serves two primary purposes: (1) to act as a benchmark for various tasks and (2) to train foundation vision or vision-language models with a focus on seafloor morphodynamic analysis. 
In contrast to existing datasets~\cite{Singh2021TheMD,Xie2022ADW,Sethuraman2024MachineLF,Walia2023OptimizedCD}, which may specialize in single machine learning tasks or offer limited data samples, our dataset provides a diverse array of seafloor mapping tasks sourced from geographically diverse regions. 
Additionally, we make our data processing source code publicly available, encouraging further expansion of the dataset towards the magnitude of large-scale natural imagery datasets~\cite{sharma-etal-2018-conceptual,changpinyo2021conceptual,jia2021scaling,schuhmann2021laion400m,schuhmann2022laion5b,liu2023visual}. 

\vspace{5pt}
\textbf{Datasets in other Scientific Domains.}
Following the success of large foundation models in natural imagery~\cite{radford2021learning,dai2023instructblip,zhu2023minigpt4,zheng2023minigpt5,liu2023visual,chen2024internvl,bai2023qwenvl,zhang2023gpt4roi,ren2023pixellm,lai2023lisa,yang2024lisa,zhang2024groundhog}, there has been a significant push to develop expansive datasets tailored for training large foundation models for specific domain applications. 
In remote sensing, initiatives such as RSVQA~\cite{lobry2020rsvqa}, RSVQA-BEN~\cite{lobry2021rsvqa}, and RSGPT~\cite{hu2023rsgpt} have been developed to enhance general VQA capabilities, while MUSE~\cite{kuckreja2023geochat} targets more complex reasoning tasks. 
Similarly, in medical imaging, datasets such as PathVQA~\cite{he2020pathvqa}, PMC-VQA~\cite{zhang2023pmcvqa}, XrayGPT~\cite{thawkar2023xraygpt}, LLaVA-Med~\cite{li2024llava}, and OmniMedVQA~\cite{hu2024omnimedvqa} aim to improve the visual and textual understanding of various body parts through the analysis of MRI, X-rays, \textit{etc.} 
These datasets comprise hundreds of millions of samples, posing significant acquisition challenges, particularly in marine science where data annotation is notably expensive. 
To address this, our initiative seeks to develop a large-scale dataset, aiming to significantly expand the resources available for marine science.\vspace{-8pt}

\section{The \texttt{SeafloorAI} Dataset} \label{sec:v}
\subsection{Dataset Overview}
\texttt{SeafloorAI} is a large, geo-distributed, multi-purpose dataset designed to map various geological layers of the seafloor. 
It is catered for training computer vision models, \textit{i.e.} CNNs and Vision Transformers that produce semantic segmentation masks.
Furthermore, it facilitates the studies of fundamental ML problems such as robust optimization~\cite{qiao2020learning,peng2022out,qiao2023topologyaware,nguyen2024adaptive}.
The dataset also serves as a basis for constructing the generative vision-language variant, \texttt{SeafloorGenAI}, discussed in Sec.~\ref{sec:vl}.

Our dataset is compiled from 62 geological data surveys published on USGS and NOAA repositories, spanning an area of 17,300 square kilometers.
This dataset features a broad geographical distribution, covering the nearshore zones of several states, including California~\cite{Golden2013-nv}, Massachusetts~\cite{Pendleton2015Nahant,Pendleton2013CapeCod,Pendleton2018Nantucket,Ackerman2015-bk}, Delmarva~\cite{Pendleton2019-ci}, and South Carolina~\cite{Baldwin2004SC}. These areas are further divided into 9 regions.
The data for this dataset were collected over a period spanning from 2004 to 2024, using a variety of single side-scan sonars and multibeam echosounders with different frequencies.
These instruments were employed to record the texture (Backscatter) and depth (Bathymetry) of the seafloor.

The surveys have been meticulously annotated by domain experts, focusing on five key geological layers: Sediment, Physiographic Zone, Habitat, Fault, and Fold as detailed in Tab.~\ref{tab:surveys}.
This expansive and detailed dataset provides a comprehensive view of geological and environmental features across a wide range of coastal environments.
In summary, we convert the raw raster data into a large-scale machine learning-ready dataset containing 696,515 input samples, and 827,220 annotated segmentation masks across various layers.

\subsection{Data Processing}

The input layers, consisting of Backscatter and Bathymetry signals, are provided as raster data in \texttt{GeoTIFF} format. 
The five mapping layers serve as the ground-truth annotations, defining five tasks for the model training and evaluation. 
These layers come in \texttt{shapefile} format that stores the location (\textit{i.e.}, longitude and latitude), shape (\textit{i.e.}, polygons) and attributes of geological features (\textit{i.e.}, sediment type).
These polygons define the regions of interest on raster images, effectively delineating the boundaries of different categories that we want to segment.

Next, we present the steps for data processing at a high level, and then go further into details with each geological layer.
First of all, we reproject all layers from all surveys to the WGS84 (EPSG:4326)\footnote{More information at \url{https://docs.up42.com/data/reference/utm}.} coordinate reference system. 
Then, we rasterize the \texttt{shapefile} to \texttt{GeoTIFF} format, effectively converting all the annotations into 2D arrays occupying the same geo-location. 
Finally, we use a sliding window to split the 2D raster layers into 224$\times$224 patches with a step size of 56 to avoid information loss at the edges.
These patches serve as the inputs and outputs for the machine learning algorithms.
This process is also referred to as ``patchifying''.

\textbf{Input Layers: Backscatter \& Bathymetry.}
Backscatter in marine science refers to the amplitude of the echoes of sound waves emitted/received by a transducer that bounce off objects or the seafloor and return to the receiver. 
By analyzing the time it takes for the sound waves to return and their acoustic intensity, scientists and researchers can create underwater maps of the submerged terrain and identify the composition and characteristics of the seafloor, as well as the presence of underwater objects or marine life. 
In our dataset, we normalize the backscatter signals to the [0, 255] range, with 255 representing the nodata value.
Regarding Bathymetry, we set the nodata value to be a negative number of significant magnitude, \textit{i.e.}, -9999.
Additionally, we convert Bathymetry measurements from meters to kilometers, compressing these values into a [0,1] range for normalization purposes.

We further calculate two morphologic derivatives from Bathymetry, namely Slope and Rugosity, to more comprehensively represent the topographical features of the seafloor in the input space. 
Slope refers to the \textit{steepness} of the seabed, calculated as the rate of change in elevation over a given distance. 
It is crucial for understanding sediment transport, habitat diversity, and the stability of underwater structures.
We use \texttt{GDAL}~\cite{gdal} implementation of the Zevenbergen \& Thorne formula~\cite{ZevenbergenThorne} to estimate the slope. In brevity, the formula computes the differences in elevation between a central pixel and its eight surrounding pixels for a more smoothed and stable slope estimation.
Rugosity, on the other hand, measures the \textit{roughness} or irregularity of the ocean floor.
It quantifies the amount of surface area relative to a flat plane, offering vital clues about the complexity of habitats, which affects biodiversity and ecological interactions. 

\begin{wrapfigure}{r}{0.35\linewidth}
\vspace{-18pt}
  \begin{center}
  \includegraphics[width=0.7\linewidth]{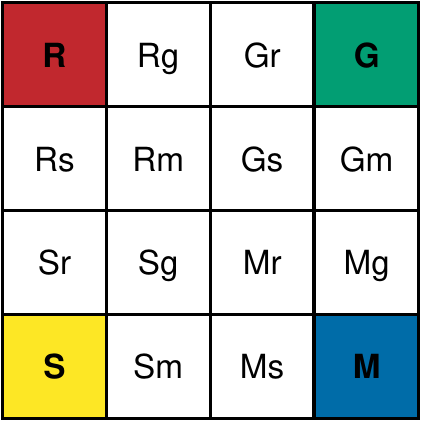}
  \end{center}
  \vspace{-5pt}
  \caption{The Barnhardt classification scheme~\cite{Barnhardt98} is based on four end-member units: (\textbf{\textcolor{redrock}{R}})ock, (\textbf{\textcolor{greengravel}{G}})ravel, (\textbf{\textcolor{yellowsand}{S}})and, and (\textbf{\textcolor{bluemud}{M}})ud. The other twelve composite categories represent the combinations of the four units, where the dominant texture ($>$ 50$\%$) is in upper case, and the subordinate ($<$ 50$\%$) in lower.}
  \label{fig:barnhardt}
\end{wrapfigure}

For each region, we resample Bathymetry, Slope and Rugosity to the Backscatter's resolution. 
As a result, our dataset contains a range of resolutions across regions, from 1m to 10m per pixel, enabling both coarse and fine-grained understanding of seafloor morphodynamic analysis.
After patchifying the rastered map, we only keep patches where the number of nodata pixels is below 10\% the number of total pixels. 
In the final step, we apply interpolation to fill in the missing pixels, and median filtering to reduce speckle noise.
The input contains 6 channels, including these 4 layers and 2 geo-location channels (pixel-wise longitude and latitude), resulting in a dimension of 224$\times$224$\times$6.


\textbf{Mapping Layers: Sediment, Physiographic Zone \& Habitat.}
Our dataset is derived from $62$ different surveys spanning both the East and West Coasts of the United States. 
Given the diverse origins of the data, there are inherent inconsistencies in the annotations, such as varying standards or differing vocabularies used to label the same categories. 
To address this, we have developed a unification process for ground-truth labels, leading to the creation of multi-class segmentation masks for Sediment, Physiographic Zone, and Habitat. 
This standardization process is meticulously overseen by domain experts to ensure the accuracy and quality of the annotations.

\textbf{(1) Sediment.}
Sediments on the seafloor, composed of varied particles from multiple sources, are crucial for creating habitats, indicating geological processes, and aiding in environmental and ecological research. They play a key role in resource exploration by helping to identify potential sites for natural resource extraction and in climate change studies by preserving historical climate data. Detailed seafloor mapping using sediment analysis is vital for accurate marine navigation, scientific research, and effective marine resource management.
We define a unified annotation standard for the Sediment layer, following the Barnhardt classification table~\cite{Barnhardt98}, which is a classification scheme based on four end-member units: (\textbf{\textcolor{redrock}{R}})ock, (\textbf{\textcolor{greengravel}{G}})ravel, (\textbf{\textcolor{yellowsand}{S}})and, and (\textbf{\textcolor{bluemud}{M}})ud. 
The other twelve composite units represent the combinations of the four units, where the dominant texture ($>$ 50$\%$ of the area) is in upper case, and the subordinate ($<$ 50$\%$ of the area) is in lower case, illustrated in Fig.~\ref{fig:barnhardt}. 
Finally, we construct semantic segmentation masks for each input patch where each pixel contains an integer value from 0 to 16, with 0 denoting the pixels without annotations.

\begin{figure}[!htb]
    \centering
    \includegraphics[width=0.85\linewidth]{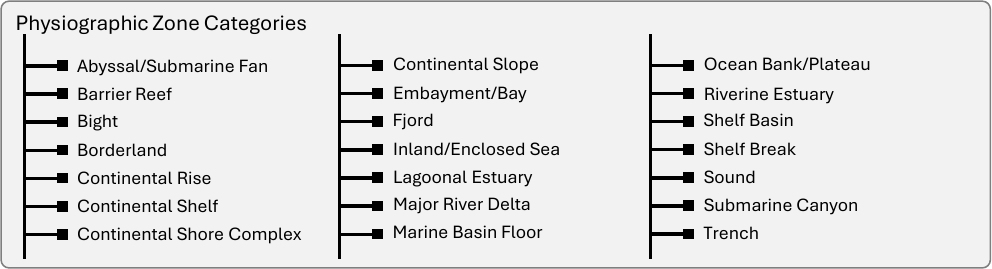}
    \caption{Twenty-one physiographic zone categories from CMECS.}
    \label{fig:pz}
\end{figure}

\textbf{(2) Physiographic Zone.} By definition, a physiographic zone refers to a distinct geographical region characterized by a uniformity in topography and underlying geological structure that sets it apart from adjacent areas. 
These zones are typically defined based on natural landscape features, such as the configuration of the terrain, rock formations, and soil types. 
Classifying these zones requires the holistic understanding of multiple geological features, hence the necessity to include the bathymetric derivatives, such as Slope and Rugosity, as input.
Similar to Sediment, we also define a standard for the Physiographic Zone layer. 
We follow the CMECS unit code for Physiographic Province which belongs in the Geoform Component~\cite{noaaCoastalMarine}. 
There are 21 different categories for Physiographic Zone, as shown in Fig.~\ref{fig:pz}. 

\begin{wrapfigure}{r}{0.35\linewidth}
\vspace{-12pt}
  \begin{center}
  \includegraphics[width=0.8\linewidth]{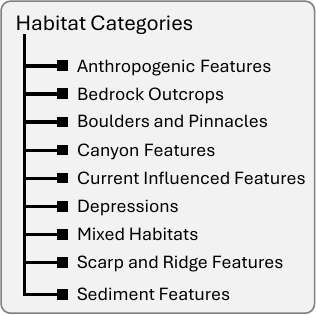}
  \end{center}
  \vspace{-5pt}
  \caption{Nine major categories for abiotic habitat defined in \texttt{SeafloorAI}.}
  \vspace{-5pt}
  \label{fig:habitats}
\end{wrapfigure}

\textbf{(3) Habitat.} One of the aims of seafloor mapping efforts is to delineate benthic habitats as a high-level outcome.
Hall et al.~\cite{Hall1997} defined Habitat as ``the resources and conditions present in an area that produce occupancy \dots by a given organism.''
According to CMECS, a benthic habitat refers to the ecological regions at the lowest level of a body of water, including the sediment surface and sub-surface layers~\cite{noaaCoastalMarine}. Benthic habitats are critical areas because they provide living space for a wide range of organisms, both flora and fauna, which are integral to the marine ecosystem.
Specifically focusing on abiotic benthic habitats, these are characterized by non-living physical and chemical aspects of the environment that influence the type and abundance of organisms living there. 
To unify the annotations across surveys, we first gather all 144 descriptions of the polygons from the public data surveys. We then categorize these descriptions into broader groups, ultimately consolidating them into 9 distinct categories for Habitat, depicted in Fig.~\ref{fig:habitats}.

\vspace{5pt}
\textbf{Mapping Layers: Fault \& Fold.}
Faults and folds are significant geological features on the seafloor that are formed by tectonic movements within the Earth's crust.
Faults occur when rock layers break and slide past each other due to tectonic forces, creating distinct disruptions in the seabed. 
Folds are bends in rock layers that occur when these layers are compressed and folded, resulting in curved or wavy stratifications. 
Detecting these features is crucial for understanding seismic activity and geological history of the marine environment.
In our study, we formulate the binary segmentation task to identify the presence of these geological features within specific image patches, assigning the pixels containing the features a value of 1, and 0 otherwise.
\vspace{-5pt}
\section{The \texttt{SeafloorGenAI} Dataset} \label{sec:vl}
\texttt{SeafloorGenAI} incorporates vision and language understanding via visual question answering (VQA), facilitating the advancement of large vision-language models in the marine science field and the conventional studies on multi-modal learning~\cite{radford2021learning,lai2023lisa,li2024llava,ma2021smil,Ma_2022_CVPR}. This integration enables smooth interactions between domain experts and AI, providing clear explanations and streamlining the process of data analysis and discovery. Our dataset, consisting of 7M QA pairs and 696K language descriptions, is designed to support \textit{general VQA capability} and \textit{instruction-following mapping}.

\begin{figure}[t!]
    \centering
    \includegraphics[width=\linewidth]{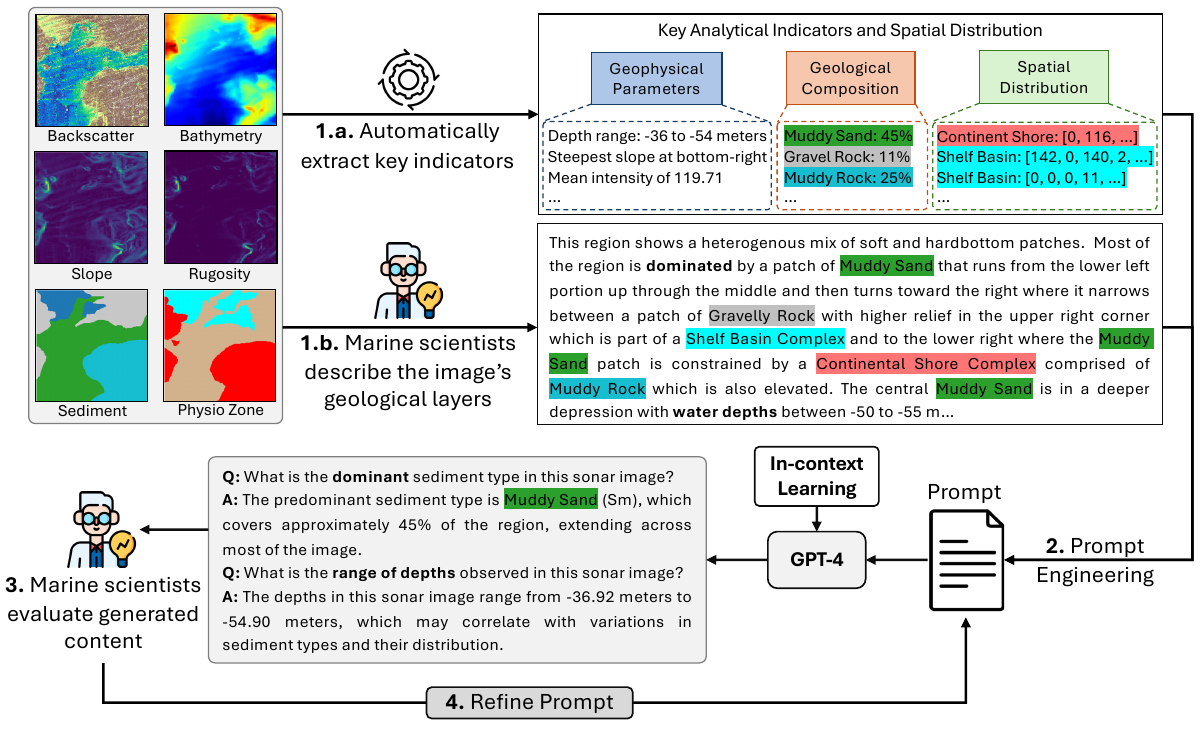}
    \caption{Pipeline for generating question-answer pairs for sonar imagery samples using \texttt{GPT-4}: Marine scientists first identify necessary information, followed by the extraction of \textit{geophysical parameters}, \textit{geological composition}, and \textit{spatial distribution}. They then provide descriptions for a handful of samples from the \texttt{SeafloorAI} dataset. These description are used to design a prompt for \texttt{GPT-4} to generate high-quality, domain-specific question-answer pairs, via in-context learning~\cite{brown2020language}.}
    \label{fig:vlm}
\end{figure}

\vspace{5pt}
\textbf{General descriptions and VQA.} Following previous work from other domains~\cite{li2024llava,kuckreja2023geochat}, we utilize large language models (LLMs), specifically \texttt{GPT-4}, to generate the language descriptions and question-answer pairs for each sonar imagery sample.
We employ in-context learning (ICL)~\cite{brown2020language}, providing few-shot input-output pairs for the LLM.
In this case, the input contains the \textit{key analytical indicators} and the output is the description written by the marine scientists for the same image.
To construct the ICL input, we, in collaboration with marine scientists, \textbf{identify} the essential information required for analysis.
Subsequently, we use standard statistical and computer vision tools to \textbf{extract} three categories of information: (1) \textit{geophysical parameters}, (2) \textit{spatial distribution} and (3) \textit{geological composition}.
The objective is to help the model ``see'' the sonar image through as much detailed language descriptions as possible.
For the ICL output, we ask marine scientists to manually \textbf{describe} in domain language 50 randomly selected samples from the \texttt{SeafloorAI} dataset.
ICL ensures \texttt{GPT-4} can accurately mimic the domain-specific language, enhancing the quality and relevance of the generated answers.
Next, we design a \textbf{prompt} to \texttt{GPT-4}, comprised of the input-output pairs and the extracted analytical indicators, to generate general descriptions and question-answer pairs for the remaining images.
Finally, the domain experts carefully \textbf{evaluate} the generated language annotations to ensure quality and consistency.
The last two steps form a feedback loop, creating an iterative prompt refinement process. 
Fig.~\ref{fig:vlm} illustrates the described pipeline.

\begin{figure}
    \centering
    \includegraphics[width=\linewidth]{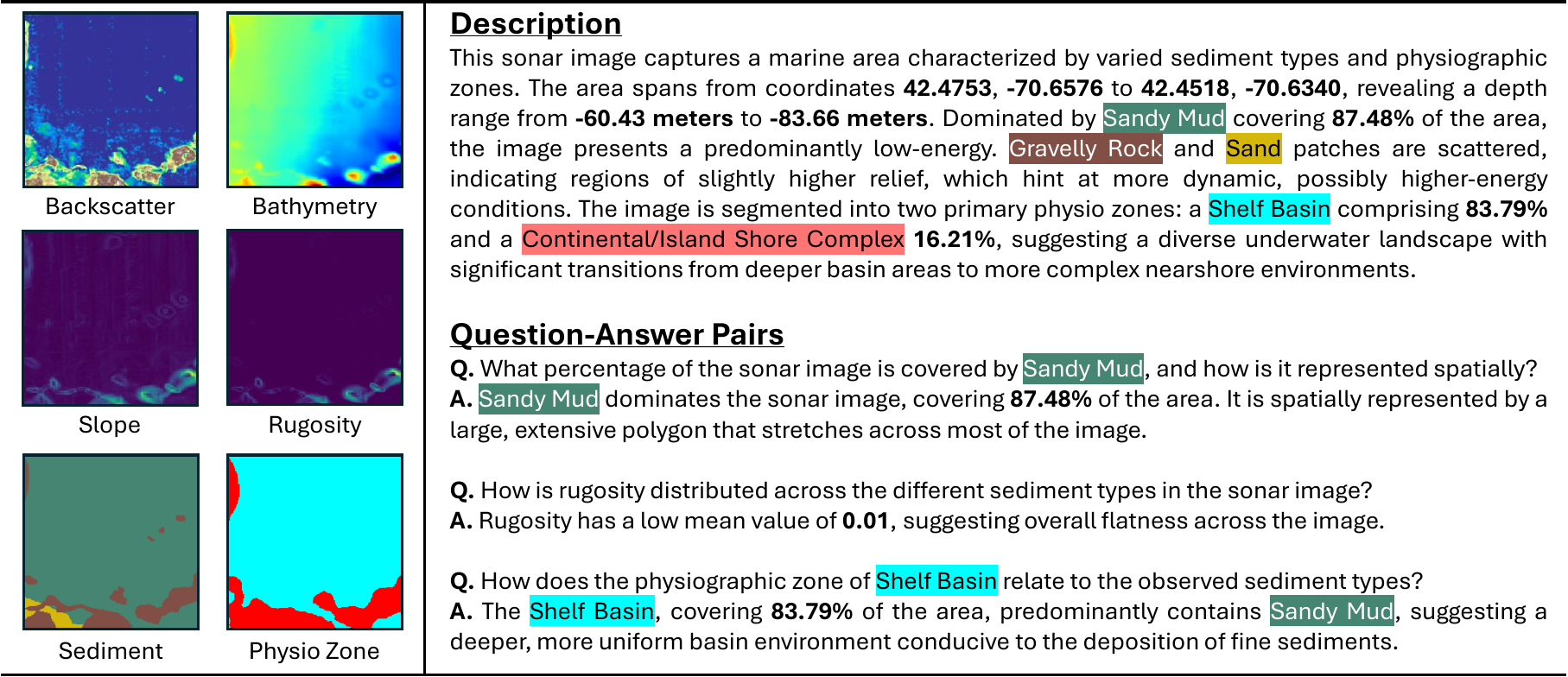}
    \caption{An example in the \texttt{SeafloorGenAI} dataset, originated from Region $5$. It features a \texttt{GPT-4} generated description and question-answer pairs designed to efficiently assist marine scientists in data analysis. The generated description covers all three key analytical indicators. Noticeably, the last QA pairs focuses on cross-layer understanding (\textit{i.e.}, Sediment and Physiographic Zone), which is helpful for unraveling complex ecological dynamics on the seabed.}
    \label{fig:demo}
    \vspace{-5pt}
\end{figure}

In Fig.~\ref{fig:demo}, we show a sample selected from the \texttt{SeafloorGenAI} dataset. We can see that \texttt{GPT-4} is able to generate QA pairs that relate different geological layers at the same location. This helps unravel complex ecological dynamics, which is beneficial to many domain applications.
We now discuss how each type of information (\textit{i.e.} geophysical parameters, spatial distribution and geological composition) is extracted from the image.


\textbf{(1) Geophysical parameters.}
These parameters are important, serving as the base for further analysis of the area. In our data processing pipeline, we employ classical analysis techniques to extract key geophysical parameters from processed data, such as water depth, mean and standard deviation of backscatter intensity, and ranges of slope, \textit{etc.}
These parameters are then systematically converted into textual format. 
This transformation facilitates a structured representation of complex numerical data, making it more accessible and interpretable for further analysis and reporting. 

\begin{tcolorbox}[breakable, colback=mycolback, colframe=myframecolor,title=An example of Geophysical Parameters in the Input layers]
\begin{Verbatim}[commandchars=\\\{\}]
Geolocation: (42.55\textdegree, -70.67\textdegree) to (42.53\textdegree, -70.64\textdegree)
Depth range: -36.4 to -54.2 meters
Backscatter mean and standard deviation: 119.7 and 72.2
Slope range: 1.7 to 9.2 degrees
Rugosity range: 0.01 to 0.02
\end{Verbatim}
\end{tcolorbox}

\vspace{5pt}
\textbf{(2) Geological composition.}
Understanding geological composition allows marine scientists to gain a holistic view of seafloor characteristics by examining how geological features are proportionally distributed within a specific area. 
Technically, this involves calculating the ratio of total pixels for each geological category relative to the overall pixels in the segmentation mask. 
As a result, we achieve the following:

\begin{tcolorbox}[breakable, colback=mycolback, colframe=myframecolor,title=An example of Geological Composition in the Sediment layer]
\begin{Verbatim}[commandchars=\\\{\}]
Muddy Sand (Sm) accounts for 45% of the image.
Muddy Rock (Rm) accounts for 25% of the image.
Gravel Rock (Rg) accounts for 11% of the image.
\end{Verbatim}
\end{tcolorbox}

\vspace{5pt}
\textbf{(3) Spatial distribution.}
Spatial distribution complements geological composition, thus giving a more comprehensive description of the image.
We convert the segmentation mask of each category to polygons, which can then be fed as language into \texttt{GPT-4}. We first find the contours of the masks using conventional computer vision techniques, then transform them into polygon representation with the format $[x_1, y_1, ..., x_n, y_n]$, where $x_i$ and $y_i$ are the coordinates of the $i^{\text{th}}$ point in $n$ points.

\begin{tcolorbox}[breakable, colback=mycolback, colframe=myframecolor,title=An example of Spatial Distribution in the Physiographic Zone layer]
\begin{Verbatim}[commandchars=\\\{\}]
Continential Shore Complex polygon at [0, 116, 0, 186, ..., 1, 117]
Shelf Basin polygon at [142, 0, 140, 2, ..., 156, 0]
\end{Verbatim}
\end{tcolorbox}

\textbf{Instruction-following Mapping.} 
Besides VQA, we aim to equip the AI assistant with the capability to map various seafloor features across different layers in response to specific instructions.
This facilitates a seamless and intuitive interaction between the AI and marine scientists, allowing for easy querying and efficient analysis.
We design our dataset to be compatible with state-of-the-art VLM models, such as PixelLM~\cite{ren2023pixellm} and LISA~\cite{lai2023lisa} for both single and multi-instance segmentation tasks. 

\begin{tcolorbox}[breakable, colback=mycolback, colframe=myframecolor,title=Examples of single and multi-instance instruction-following mapping in SeafloorGenAI]
\begin{Verbatim}[commandchars=\\\{\}]
(1) Q: Please segment [CATEGORY] in [LAYER].
    A: Sure, <SEG>.

(2) Q: What are present in the image for [LAYER]? Please segment them.
    A: [CATEGORY_1] <SEG_1>, [CATEGORY_2] <SEG_2>, ..., [CATEGORY_N] <SEG_N>.

(3) Q: Identify the areas of [CATEGORY_1] from [LAYER_1] and [CATEGORY_2] 
       from [LAYER_2].
    A: Sure, [CATEGORY_1] from [LAYER_1] <SEG_1> and [CATEGORY_2] 
       from [LAYER_2] <SEG_2>.
\end{Verbatim}
\end{tcolorbox}

\section{Experiments}
We report some baseline experiment runs on \texttt{SeafloorAI} for multi-class segmentation. Due to space limit, we move the experiments for binary segmentation to the Supplementary Material.

\textbf{Evaluation Metrics.}
We use pixel-wise accuracy (Acc), Dice coefficient (Dice) and Jaccard coefficient (mIoU) to evaluate the baseline models.

\textbf{Data Split.}
We present the data splits for Sediment, Physiographic Zone and Habitat, as well as the motivation for such splits.
Due to the availability of the categories in each region, we make sure that the training regions possess the set of categories that cover the testing region(s).
We present our data splits for the layers in Tab.~\ref{tab:splits}. For the source data, we randomly split them into 90$\%$ for training and 10$\%$ for validation. The validation set is used to select the best model for testing on the target data.

\begin{table}[!htb]
\resizebox{0.8\textwidth}{!}{
\begin{tabular}{l|l|l|l}
\toprule
\textbf{Task} & \textbf{Layer} & \multicolumn{1}{c|}{\textbf{Source}}    & \multicolumn{1}{c}{\textbf{Target}} \\
\midrule
\multirow{3}{*}{\shortstack[l]{Multi-class\\ Segmentation}} & Sediment & Region 1, Region 5, Region 6, Region 7 & Region 3 \\
& Physio Zone & Region 1, Region 3, Region 5, Region 6 & Region 7 \\
& Habitat & Region 2 & Region 4 \\   
\bottomrule
\end{tabular}}
\caption{Geo-distributed data splits for the \texttt{SeafloorAI} dataset for multi-class segmentation.}
\label{tab:splits}
\end{table}

\textbf{Training Details.}
We employ the UNet architecture with different backbones as baselines. The UNet architecture~\cite{ronneberger2015u} consists of a contracting path (encoder) and an expanding path (decoder), forming a U-shape. 
We use UNet-Base~\cite{ronneberger2015u}, UNet-ResNet18~\cite{he2015deep} and TransUNet-ViT-B/32~\cite{chen2021transunet,dosovitskiy2021image} as our baseline models for the multi-class segmentation tasks.
We adopt cross-entropy as the loss function.
The model was trained using the Adam optimizer~\cite{kingma2017adam}. The learning rate was initially set to 0.001 with a cosine annealing schedule.
We use a batch size of 64 for 100 epochs, setting the patience to 5 epochs for early stopping.
We perform 3 runs with different random seeds and report the model performance in Tab.~\ref{tab:results1}.
All runs are conducted on a single NVIDIA RTX A6000 GPU.

\textbf{Results.} Tab.~\ref{tab:results1} reports the results on the geo-distributed setting, which is similar to out-of-distribution generalization~\cite{qiao2020learning, peng2022out}. We report the in-distribution (ID; on source data) and out-of-distribution (OOD; on target data) pixel-wise accuracy, Dice coefficient and Jaccard coefficient. Overall, we can see that all baseline models suffer from a signification performance degradation under distribution shift. This might be due to covariate shift (sensor types and configurations) and subpopulation shift (class imbalance). Therefore, ensuring that a model generalizes well to new, unseen distributions is a fundamental challenge. Standard training methods often assume that the training and testing data come from the same distribution, which is rarely the case in real-world applications.

\begin{table}[htb!]
\resizebox{\linewidth}{!}{
\begin{tabular}{l|ccc|ccc|ccc}
\toprule
\multirow{2}{*}{} & \multicolumn{9}{c}{\textbf{Sediment}} \\
\cmidrule{2-10}
& Acc ID & Acc OOD & $\Delta$ Acc & Dice ID & Dice OOD & $\Delta$ Dice & mIoU ID & mIoU OOD & $\Delta$ mIoU \\
\midrule
UNet-Base & 77.45 $\pm$ 0.81 & 21.49 $\pm$ 0.91 & -55.96 & 79.73 $\pm$ 0.83 & 21.59 $\pm$ 0.97 & -58.14 & 66.46 $\pm$ 1.15 & 12.29 $\pm$ 0.61 & -54.17\\
UNet-ResNet18 & 78.45 $\pm$ 0.67 & 34.71 $\pm$ 6.79 & -43.74 & 80.78 $\pm$ 0.71 & 35.01 $\pm$ 6.86 & -45.77 & 67.90 $\pm$ 1.00 & 22.08 $\pm$ 5.73 & -45.82 \\
TransUNet & 67.90 $\pm$ 2.18 & 28.32 $\pm$ 1.04 & -39.58 & 69.94 $\pm$ 2.27 & 29.16 $\pm$ 1.05 & -40.16 & 53.98 $\pm$ 2.65 & 17.41 $\pm$ 0.71 & -36.57 \\

\midrule
\multirow{2}{*}{} & \multicolumn{9}{c}{\textbf{Physio Zone}} \\
\cmidrule{2-10}
& Acc ID & Acc OOD & $\Delta$ Acc & Dice ID & Dice OOD & $\Delta$ Dice & mIoU ID & mIoU OOD & $\Delta$ mIoU \\
\midrule
UNet-Base & 93.05 $\pm$ 0.16 & 56.56 $\pm$ 0.87 & -36.49 & 95.81 $\pm$ 0.18 & 57.09 $\pm$ 0.69 & -38.72 & 91.98 $\pm$ 0.32 & 43.22 $\pm$ 0.84 & -48.76 \\
UNet-ResNet18 & 92.87 $\pm$ 0.10 & 56.74 $\pm$ 2.53 & -36.13 & 95.63 $\pm$ 0.09 & 59.86 $\pm$ 2.54 & -35.77 & 91.66 $\pm$ 0.17 & 42.97 $\pm$ 3.00 & -48.69\\
TransUNet & 90.63 $\pm$ 0.20 & 56.24 $\pm$ 1.66 & -34.39 & 93.28 $\pm$ 0.27 & 57.51 $\pm$ 1.84 & -35.77 & 87.49 $\pm$ 0.47 & 43.86 $\pm$ 2.04 & -43.63\\

\midrule
\multirow{2}{*}{} & \multicolumn{9}{c}{\textbf{Habitat}}  \\
\cmidrule{2-10}
& Acc ID & Acc OOD & $\Delta$ Acc & Dice ID & Dice OOD & $\Delta$ Dice & mIoU ID & mIoU OOD & $\Delta$ mIoU \\
\midrule
UNet-Base & 92.02 $\pm$ 0.18 & 70.54 $\pm$ 1.72 & -21.48 & 94.82 $\pm$ 0.20 & 71.04 $\pm$ 1.54 & -23.78 & 90.19 $\pm$ 0.37 & 56.75 $\pm$ 2.01 & -33.44\\
UNet-ResNet18 & 92.70 $\pm$ 0.12 & 76.40 $\pm$ 1.33 & -16.30 & 95.50 $\pm$ 0.11 & 76.59 $\pm$ 1.28 & -18.91 & 91.43 $\pm$ 0.20 & 65.17 $\pm$ 1.80 & -26.26\\
TransUNet & 88.67 $\pm$ 0.56 & 70.56 $\pm$ 0.72 & -18.11 & 91.34 $\pm$ 0.59 & 72.76 $\pm$ 0.83 & -18.58 & 84.15 $\pm$ 0.99 & 59.38 $\pm$ 1.24 & -24.77\\

\bottomrule
\end{tabular}
}
\caption{Performance of the baselines in the geo-distributed setting for multi-class segmentation.}
\label{tab:results1}
\end{table}
\section{Human Evaluation for Language Annotations}
Although \texttt{GPT-4} has shown strong capabilities in data annotations~\cite{tan2024largelanguagemodelsdata}, hallucinations in LLMs are inevitable~\cite{huang2023surveyhallucinationlargelanguage}. To ensure the quality of the language annotations generated by \texttt{GPT-4}, we describe an iterative prompt refinement process that involves human expert evaluation.

\begin{wrapfigure}{r}{0.48\linewidth}
\vspace{-20pt}
  \begin{center}
  \includegraphics[width=\linewidth]{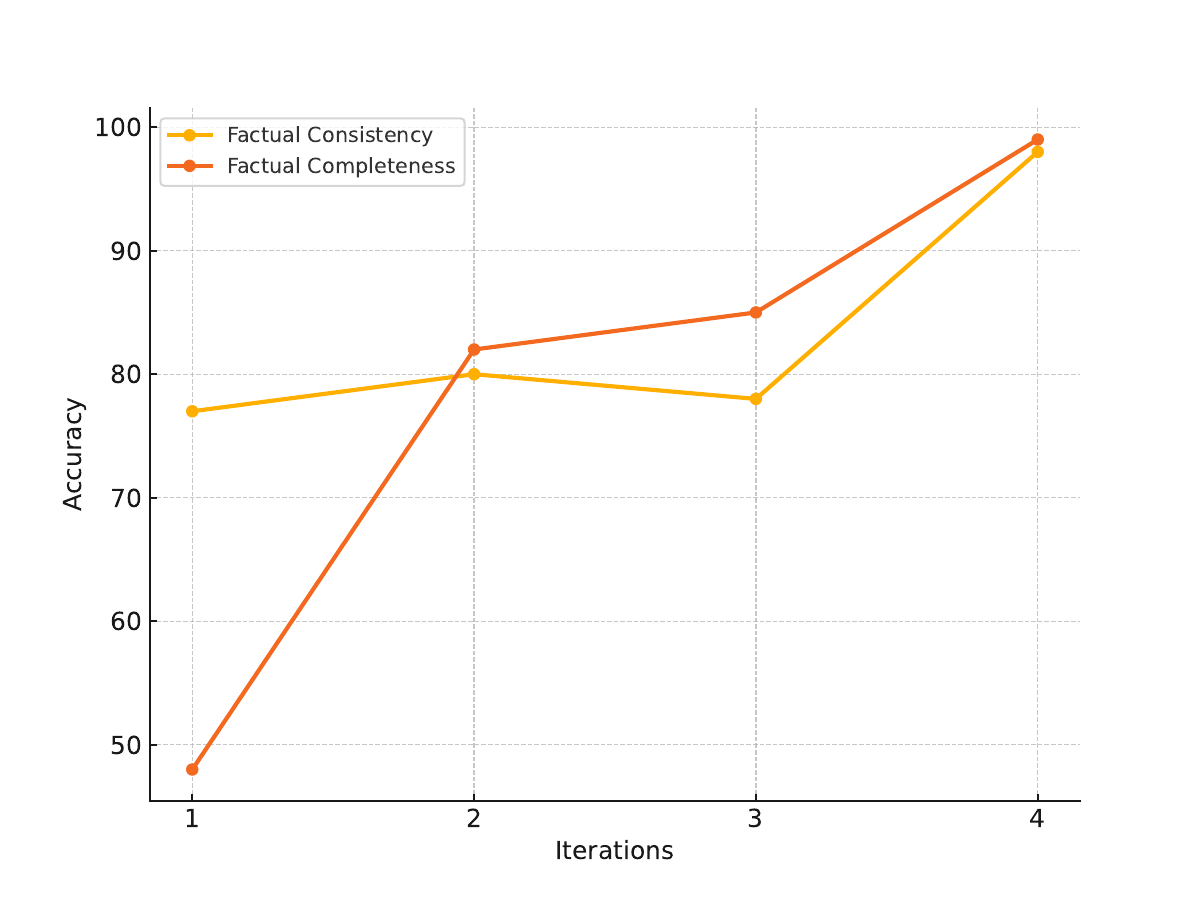}
  \end{center}
  \vspace{-5pt}
  \caption{Accuracy for factual consistency and completeness increases over the iterations thanks to rigorous the prompt refinement procedure. \texttt{GPT-4} performs worse on factual completeness potentially due to hallucinations.}
  \label{fig:humaneval}
\end{wrapfigure}
To maximize budget efficiency, we designed our procedure with several iterations of feedback and refinement. The idea is to engineer and refine our prompt to \texttt{GPT-4} on a small subset of data before applying it to the whole dataset. For each iteration, (1) we annotated 1,000 random samples with \texttt{GPT-4}; (2) the marine scientists reviewed the quality of the generated annotations and gave feedback based on the three criteria: (i) \textit{factual consistency} to the original annotations, (ii) \textit{factual completeness} with respect to the analytical indicators and (iii) \textit{coherence} to domain language; (3) we refined our prompts to \texttt{GPT-4}, a.k.a prompt engineering, to achieve higher quality language annotations, (4) we repeated the steps for the next iteration. Finally, when the quality is met, we will populate the entire dataset with language annotations. Due to space limitation, we include more details in the Supplementary Material.

\vspace{-5pt}
\section{Limitations and Future Work}
Despite the extensiveness of our dataset, there are notable limitations to discuss. Firstly, the availability of layers is not uniform across all regions; for instance, Region $5$ is missing Habitat, Fault, and Fold layers. This is due to the different mapping objectives when the data surveys were first collected. Additionally, the existing nine Habitat categories are somewhat coarse and exclude biotic classifications. We are actively collaborating with marine scientists to refine and expand the Habitat layer, making it more detailed and comprehensive.
The current version of the \texttt{SeafloorGenAI} dataset provides annotations suitable for straightforward analytical queries and lacks the data for deeper reasoning abilities. Moving forward, we plan to enhance the dataset to support the development of reasoning-capable models similar to referring and reasoning segmentation as in~\cite{ren2023pixellm, lai2023lisa,yang2024lisa}, offering more profound insights into marine science questions and paving the way for data discovery.
Developing this enhanced version of the dataset will require a structured and systematic approach to understanding domain-specific knowledge to accurately annotate the data.
In terms of modeling, our plan for future work involves training a generative vision-language model on the \texttt{SeafloorGenAI} dataset, serving as a foundation ML model in marine science research. 


\section*{Acknowledgement}
This work is supported by the DoD DEPSCoR Award AFOSR FA9550-23-1-0494, the  NSF CAREER Award No. 2340074, the NSF SLES Award No. 2416937, and the NSF III CORE Award No. 2412675.
Any opinions, findings and conclusions or recommendations expressed in this material are
those of the authors and do not reflect the views of the supporting entities.

{
\small
\bibliographystyle{plain}
\bibliography{neurips_2024}

\begin{thebibliography}{10}

\bibitem{noaaCoastalMarine}
{C}oastal and {M}arine {E}cological {C}lassification {S}tandard ({C}{M}{E}{C}{S}) --- repository.library.noaa.gov.
\newblock \url{https://repository.library.noaa.gov/view/noaa/27552}.
\newblock [Accessed 24-05-2024].

\bibitem{Ackerman2015-bk}
Seth~D Ackerman, Adrienne~L Pappal, Emily~C Huntley, Dann~S Blackwood, and William~C Schwab.
\newblock Geological sampling data and benthic biota classification: Buzzards bay and vineyard sound, massachusetts, 2015.

\bibitem{arosio2023fully}
Riccardo Arosio, Brandon Hobley, Andrew~J Wheeler, Fabio Sacchetti, Luis~A Conti, Thomas Furey, and Aaron Lim.
\newblock Fully convolutional neural networks applied to large-scale marine morphology mapping.
\newblock {\em Frontiers in Marine Science}, 10:1228867, 2023.

\bibitem{bai2023qwenvl}
Jinze Bai, Shuai Bai, Shusheng Yang, Shijie Wang, Sinan Tan, Peng Wang, Junyang Lin, Chang Zhou, and Jingren Zhou.
\newblock Qwen-vl: A versatile vision-language model for understanding, localization, text reading, and beyond, 2023.

\bibitem{Baldwin2004SC}
Wayne~E. Baldwin, Robert~A. Morton, Jane~F. Denny, William C.~Schwab Shawn V.~Dadisman, Paul~T. Gayes, and Neal~W. Driscoll.
\newblock Maps showing the stratigraphic framework of south carolina's long bay from little river to winyah bay, 2004.

\bibitem{Barnhardt98}
Walter~A. Barnhardt, Joseph~T. Kelley, Stephen~M. Dickson, and Daniel~F. Belknap.
\newblock Mapping the gulf of maine with side-scan sonar: A new bottom-type classification for complex seafloors.
\newblock {\em Journal of Coastal Research}, 14(2):646--659, 1998.

\bibitem{berthold2017}
Tim Berthold, Artem Leichter, Bodo Rosenhahn, Volker Berkhahn, and Jennifer Valerius.
\newblock Seabed sediment classification of side-scan sonar data using convolutional neural networks.
\newblock In {\em 2017 IEEE Symposium Series on Computational Intelligence (SSCI)}, pages 1--8, 2017.

\bibitem{brown2020language}
Tom~B. Brown, Benjamin Mann, Nick Ryder, Melanie Subbiah, Jared Kaplan, Prafulla Dhariwal, Arvind Neelakantan, Pranav Shyam, Girish Sastry, Amanda Askell, Sandhini Agarwal, Ariel Herbert-Voss, Gretchen Krueger, Tom Henighan, Rewon Child, Aditya Ramesh, Daniel~M. Ziegler, Jeffrey Wu, Clemens Winter, Christopher Hesse, Mark Chen, Eric Sigler, Mateusz Litwin, Scott Gray, Benjamin Chess, Jack Clark, Christopher Berner, Sam McCandlish, Alec Radford, Ilya Sutskever, and Dario Amodei.
\newblock Language models are few-shot learners, 2020.

\bibitem{changpinyo2021conceptual}
Soravit Changpinyo, Piyush Sharma, Nan Ding, and Radu Soricut.
\newblock Conceptual 12m: Pushing web-scale image-text pre-training to recognize long-tail visual concepts, 2021.

\bibitem{chen2021transunet}
Jieneng Chen, Yongyi Lu, Qihang Yu, Xiangde Luo, Ehsan Adeli, Yan Wang, Le~Lu, Alan~L. Yuille, and Yuyin Zhou.
\newblock Transunet: Transformers make strong encoders for medical image segmentation, 2021.

\bibitem{chen2016deep}
Johnny~L Chen and Jason~E Summers.
\newblock Deep neural networks for learning classification features and generative models from synthetic aperture sonar big data.
\newblock In {\em Proceedings of Meetings on Acoustics}, volume~29. AIP Publishing, 2016.

\bibitem{chen2024internvl}
Zhe Chen, Jiannan Wu, Wenhai Wang, Weijie Su, Guo Chen, Sen Xing, Muyan Zhong, Qinglong Zhang, Xizhou Zhu, Lewei Lu, Bin Li, Ping Luo, Tong Lu, Yu~Qiao, and Jifeng Dai.
\newblock Internvl: Scaling up vision foundation models and aligning for generic visual-linguistic tasks, 2024.

\bibitem{cheng2022multi}
Zhen Cheng, Guanying Huo, and Haisen Li.
\newblock A multi-domain collaborative transfer learning method with multi-scale repeated attention mechanism for underwater side-scan sonar image classification.
\newblock {\em Remote Sensing}, 14(2):355, 2022.

\bibitem{dai2023instructblip}
Wenliang Dai, Junnan Li, Dongxu Li, Anthony Meng~Huat Tiong, Junqi Zhao, Weisheng Wang, Boyang Li, Pascale Fung, and Steven Hoi.
\newblock Instructblip: Towards general-purpose vision-language models with instruction tuning, 2023.

\bibitem{dosovitskiy2021image}
Alexey Dosovitskiy, Lucas Beyer, Alexander Kolesnikov, Dirk Weissenborn, Xiaohua Zhai, Thomas Unterthiner, Mostafa Dehghani, Matthias Minderer, Georg Heigold, Sylvain Gelly, Jakob Uszkoreit, and Neil Houlsby.
\newblock An image is worth 16x16 words: Transformers for image recognition at scale, 2021.

\bibitem{gdal}
{GDAL/OGR contributors}.
\newblock {\em {GDAL/OGR} Geospatial Data Abstraction software Library}.
\newblock Open Source Geospatial Foundation, 2024.

\bibitem{Golden2013-kn}
Nadine~E Golden.
\newblock California state waters map series data catalog, 2013.

\bibitem{Golden2013-nv}
Nadine~E Golden.
\newblock California state waters map series data catalog, 2013.

\bibitem{gui2024surveyselfsupervisedlearningalgorithms}
Jie Gui, Tuo Chen, Jing Zhang, Qiong Cao, Zhenan Sun, Hao Luo, and Dacheng Tao.
\newblock A survey on self-supervised learning: Algorithms, applications, and future trends, 2024.

\bibitem{Hall1997}
Linnea~S. Hall, Paul~R. Krausman, and Michael~L. Morrison.
\newblock The habitat concept and a plea for standard terminology.
\newblock {\em Wildlife Society Bulletin (1973-2006)}, 25(1):173--182, 1997.

\bibitem{he2015deep}
Kaiming He, Xiangyu Zhang, Shaoqing Ren, and Jian Sun.
\newblock Deep residual learning for image recognition, 2015.

\bibitem{he2020pathvqa}
Xuehai He, Yichen Zhang, Luntian Mou, Eric Xing, and Pengtao Xie.
\newblock Pathvqa: 30000+ questions for medical visual question answering, 2020.

\bibitem{hu2023rsgpt}
Yuan Hu, Jianlong Yuan, Congcong Wen, Xiaonan Lu, and Xiang Li.
\newblock Rsgpt: A remote sensing vision language model and benchmark.
\newblock {\em arXiv preprint arXiv:2307.15266}, 2023.

\bibitem{hu2024omnimedvqa}
Yutao Hu, Tianbin Li, Quanfeng Lu, Wenqi Shao, Junjun He, Yu~Qiao, and Ping Luo.
\newblock Omnimedvqa: A new large-scale comprehensive evaluation benchmark for medical lvlm, 2024.

\bibitem{huang2023surveyhallucinationlargelanguage}
Lei Huang, Weijiang Yu, Weitao Ma, Weihong Zhong, Zhangyin Feng, Haotian Wang, Qianglong Chen, Weihua Peng, Xiaocheng Feng, Bing Qin, and Ting Liu.
\newblock A survey on hallucination in large language models: Principles, taxonomy, challenges, and open questions, 2023.

\bibitem{huo2020ieee}
Guanying Huo, Ziyin Wu, and Jiabiao Li.
\newblock Underwater object classification in sidescan sonar images using deep transfer learning and semisynthetic training data.
\newblock {\em IEEE Access}, 8:47407--47418, 2020.

\bibitem{Islam2020Semantic}
Md~Jahidul Islam, Chelsey Edge, Yuyang Xiao, Peigen Luo, Muntaqim Mehtaz, Christopher Morse, Sadman~Sakib Enan, and Junaed Sattar.
\newblock Semantic segmentation of underwater imagery: Dataset and benchmark.
\newblock In {\em 2020 IEEE/RSJ International Conference on Intelligent Robots and Systems (IROS)}, pages 1769--1776, 2020.

\bibitem{jia2021scaling}
Chao Jia, Yinfei Yang, Ye~Xia, Yi-Ting Chen, Zarana Parekh, Hieu Pham, Quoc~V. Le, Yunhsuan Sung, Zhen Li, and Tom Duerig.
\newblock Scaling up visual and vision-language representation learning with noisy text supervision, 2021.

\bibitem{kingma2017adam}
Diederik~P. Kingma and Jimmy Ba.
\newblock Adam: A method for stochastic optimization, 2017.

\bibitem{kuckreja2023geochat}
Kartik Kuckreja, Muhammad~S. Danish, Muzammal Naseer, Abhijit Das, Salman Khan, and Fahad~S. Khan.
\newblock Geochat: Grounded large vision-language model for remote sensing.
\newblock {\em The IEEE/CVF Conference on Computer Vision and Pattern Recognition}, 2024.

\bibitem{lai2023lisa}
Xin Lai, Zhuotao Tian, Yukang Chen, Yanwei Li, Yuhui Yuan, Shu Liu, and Jiaya Jia.
\newblock Lisa: Reasoning segmentation via large language model.
\newblock {\em arXiv preprint arXiv:2308.00692}, 2023.

\bibitem{li2021zero}
Chuanlong Li, Xiufen Ye, Dongxiang Cao, Jie Hou, and Haibo Yang.
\newblock Zero shot objects classification method of side scan sonar image based on synthesis of pseudo samples.
\newblock {\em Applied Acoustics}, 173:107691, 2021.

\bibitem{li2024llava}
Chunyuan Li, Cliff Wong, Sheng Zhang, Naoto Usuyama, Haotian Liu, Jianwei Yang, Tristan Naumann, Hoifung Poon, and Jianfeng Gao.
\newblock Llava-med: Training a large language-and-vision assistant for biomedicine in one day.
\newblock {\em Advances in Neural Information Processing Systems}, 36, 2024.

\bibitem{li2024deal}
Tang Li, Mengmeng Ma, and Xi~Peng.
\newblock Deal: Disentangle and localize concept-level explanations for vlms.
\newblock In {\em European Conference on Computer Vision}, pages 383--401. Springer, 2025.

\bibitem{liu2023visual}
Haotian Liu, Chunyuan Li, Qingyang Wu, and Yong~Jae Lee.
\newblock Visual instruction tuning.
\newblock In {\em Thirty-seventh Conference on Neural Information Processing Systems}, 2023.

\bibitem{lobry2021rsvqa}
Sylvain Lobry, Beg{\"u}m Demir, and Devis Tuia.
\newblock Rsvqa meets bigearthnet: a new, large-scale, visual question answering dataset for remote sensing.
\newblock In {\em 2021 IEEE International Geoscience and Remote Sensing Symposium IGARSS}, pages 1218--1221. IEEE, 2021.

\bibitem{lobry2020rsvqa}
Sylvain Lobry, Diego Marcos, Jesse Murray, and Devis Tuia.
\newblock Rsvqa: Visual question answering for remote sensing data.
\newblock {\em IEEE Transactions on Geoscience and Remote Sensing}, 58(12):8555--8566, 2020.

\bibitem{lundine2023deep}
Mark~A Lundine, Laura~L Brothers, and Arthur~C Trembanis.
\newblock Deep learning for pockmark detection: Implications for quantitative seafloor characterization.
\newblock {\em Geomorphology}, 421:108524, 2023.

\bibitem{luo2019ieee}
Xiaowen Luo, Xiaoming Qin, Ziyin Wu, Fanlin Yang, Mingwei Wang, and Jihong Shang.
\newblock Sediment classification of small-size seabed acoustic images using convolutional neural networks.
\newblock {\em IEEE Access}, 7:98331--98339, 2019.

\bibitem{Ma_2022_CVPR}
Mengmeng Ma, Jian Ren, Long Zhao, Davide Testuggine, and Xi~Peng.
\newblock Are multimodal transformers robust to missing modality?
\newblock In {\em Proceedings of the IEEE/CVF Conference on Computer Vision and Pattern Recognition (CVPR)}, pages 18177--18186, June 2022.

\bibitem{ma2021smil}
Mengmeng Ma, Jian Ren, Long Zhao, Sergey Tulyakov, Cathy Wu, and Xi~Peng.
\newblock Smil: Multimodal learning with severely missing modality.
\newblock In {\em Proceedings of the AAAI Conference on Artificial Intelligence}, volume~35, pages 2302--2310, 2021.

\bibitem{Mayer2018-cx}
Larry Mayer, Martin Jakobsson, Graham Allen, Boris Dorschel, Robin Falconer, Vicki Ferrini, Geoffroy Lamarche, Helen Snaith, and Pauline Weatherall.
\newblock The nippon {foundation---GEBCO} seabed 2030 project: The quest to see the world's oceans completely mapped by 2030.
\newblock {\em Geosciences (Basel)}, 8(2):63, February 2018.

\bibitem{mckay2017s}
John McKay, Isaac Gerg, Vishal Monga, and Raghu~G Raj.
\newblock What's mine is yours: Pretrained cnns for limited training sonar atr.
\newblock In {\em OCEANS 2017-anchorage}, pages 1--7. IEEE, 2017.

\bibitem{nayak2021machine}
Nandeeka Nayak, Makoto Nara, Timmy Gambin, Zo{\"e} Wood, and Christopher~M Clark.
\newblock Machine learning techniques for auv side-scan sonar data feature extraction as applied to intelligent search for underwater archaeological sites.
\newblock In {\em Field and Service Robotics: Results of the 12th International Conference}, pages 219--233. Springer, 2021.

\bibitem{nguyen2024adaptive}
Kien~X. Nguyen, Fengchun Qiao, and Xi~Peng.
\newblock Adaptive cascading network for continual test-time adaptation.
\newblock In {\em Proceedings of the 33rd ACM International Conference on Information and Knowledge Management}, CIKM '24, page 1763–1773, New York, NY, USA, 2024. Association for Computing Machinery.

\bibitem{openai2024gpt4}
OpenAI, Josh Achiam, Steven Adler, Sandhini Agarwal, Lama Ahmad, Ilge Akkaya, Florencia~Leoni Aleman, Diogo Almeida, Janko Altenschmidt, Sam Altman, Shyamal Anadkat, Red Avila, Igor Babuschkin, Suchir Balaji, Valerie Balcom, Paul Baltescu, Haiming Bao, Mohammad Bavarian, Jeff Belgum, Irwan Bello, Jake Berdine, Gabriel Bernadett-Shapiro, Christopher Berner, Lenny Bogdonoff, Oleg Boiko, Madelaine Boyd, Anna-Luisa Brakman, Greg Brockman, Tim Brooks, Miles Brundage, Kevin Button, Trevor Cai, Rosie Campbell, Andrew Cann, Brittany Carey, Chelsea Carlson, Rory Carmichael, Brooke Chan, Che Chang, Fotis Chantzis, Derek Chen, Sully Chen, Ruby Chen, Jason Chen, Mark Chen, Ben Chess, Chester Cho, Casey Chu, Hyung~Won Chung, Dave Cummings, Jeremiah Currier, Yunxing Dai, Cory Decareaux, Thomas Degry, Noah Deutsch, Damien Deville, Arka Dhar, David Dohan, Steve Dowling, Sheila Dunning, Adrien Ecoffet, Atty Eleti, Tyna Eloundou, David Farhi, Liam Fedus, Niko Felix, Simón~Posada Fishman, Juston Forte, Isabella Fulford, Leo
  Gao, Elie Georges, Christian Gibson, Vik Goel, Tarun Gogineni, Gabriel Goh, Rapha Gontijo-Lopes, Jonathan Gordon, Morgan Grafstein, Scott Gray, Ryan Greene, Joshua Gross, Shixiang~Shane Gu, Yufei Guo, Chris Hallacy, Jesse Han, Jeff Harris, Yuchen He, Mike Heaton, Johannes Heidecke, Chris Hesse, Alan Hickey, Wade Hickey, Peter Hoeschele, Brandon Houghton, Kenny Hsu, Shengli Hu, Xin Hu, Joost Huizinga, Shantanu Jain, Shawn Jain, Joanne Jang, Angela Jiang, Roger Jiang, Haozhun Jin, Denny Jin, Shino Jomoto, Billie Jonn, Heewoo Jun, Tomer Kaftan, Łukasz Kaiser, Ali Kamali, Ingmar Kanitscheider, Nitish~Shirish Keskar, Tabarak Khan, Logan Kilpatrick, Jong~Wook Kim, Christina Kim, Yongjik Kim, Jan~Hendrik Kirchner, Jamie Kiros, Matt Knight, Daniel Kokotajlo, Łukasz Kondraciuk, Andrew Kondrich, Aris Konstantinidis, Kyle Kosic, Gretchen Krueger, Vishal Kuo, Michael Lampe, Ikai Lan, Teddy Lee, Jan Leike, Jade Leung, Daniel Levy, Chak~Ming Li, Rachel Lim, Molly Lin, Stephanie Lin, Mateusz Litwin, Theresa Lopez, Ryan
  Lowe, Patricia Lue, Anna Makanju, Kim Malfacini, Sam Manning, Todor Markov, Yaniv Markovski, Bianca Martin, Katie Mayer, Andrew Mayne, Bob McGrew, Scott~Mayer McKinney, Christine McLeavey, Paul McMillan, Jake McNeil, David Medina, Aalok Mehta, Jacob Menick, Luke Metz, Andrey Mishchenko, Pamela Mishkin, Vinnie Monaco, Evan Morikawa, Daniel Mossing, Tong Mu, Mira Murati, Oleg Murk, David Mély, Ashvin Nair, Reiichiro Nakano, Rajeev Nayak, Arvind Neelakantan, Richard Ngo, Hyeonwoo Noh, Long Ouyang, Cullen O'Keefe, Jakub Pachocki, Alex Paino, Joe Palermo, Ashley Pantuliano, Giambattista Parascandolo, Joel Parish, Emy Parparita, Alex Passos, Mikhail Pavlov, Andrew Peng, Adam Perelman, Filipe de~Avila Belbute~Peres, Michael Petrov, Henrique~Ponde de~Oliveira~Pinto, Michael, Pokorny, Michelle Pokrass, Vitchyr~H. Pong, Tolly Powell, Alethea Power, Boris Power, Elizabeth Proehl, Raul Puri, Alec Radford, Jack Rae, Aditya Ramesh, Cameron Raymond, Francis Real, Kendra Rimbach, Carl Ross, Bob Rotsted, Henri Roussez,
  Nick Ryder, Mario Saltarelli, Ted Sanders, Shibani Santurkar, Girish Sastry, Heather Schmidt, David Schnurr, John Schulman, Daniel Selsam, Kyla Sheppard, Toki Sherbakov, Jessica Shieh, Sarah Shoker, Pranav Shyam, Szymon Sidor, Eric Sigler, Maddie Simens, Jordan Sitkin, Katarina Slama, Ian Sohl, Benjamin Sokolowsky, Yang Song, Natalie Staudacher, Felipe~Petroski Such, Natalie Summers, Ilya Sutskever, Jie Tang, Nikolas Tezak, Madeleine~B. Thompson, Phil Tillet, Amin Tootoonchian, Elizabeth Tseng, Preston Tuggle, Nick Turley, Jerry Tworek, Juan Felipe~Cerón Uribe, Andrea Vallone, Arun Vijayvergiya, Chelsea Voss, Carroll Wainwright, Justin~Jay Wang, Alvin Wang, Ben Wang, Jonathan Ward, Jason Wei, CJ~Weinmann, Akila Welihinda, Peter Welinder, Jiayi Weng, Lilian Weng, Matt Wiethoff, Dave Willner, Clemens Winter, Samuel Wolrich, Hannah Wong, Lauren Workman, Sherwin Wu, Jeff Wu, Michael Wu, Kai Xiao, Tao Xu, Sarah Yoo, Kevin Yu, Qiming Yuan, Wojciech Zaremba, Rowan Zellers, Chong Zhang, Marvin Zhang, Shengjia
  Zhao, Tianhao Zheng, Juntang Zhuang, William Zhuk, and Barret Zoph.
\newblock Gpt-4 technical report, 2024.

\bibitem{Pendleton2013CapeCod}
Elizabeth~A Pendleton, Wayne~E Baldwin, Walter~A Barnhardt, Seth~D Ackerman, David~S Foster, Brian~D Andrews, and William~C Schwab.
\newblock Shallow geology, seafloor texture, and physiographic zones of the inner continental shelf from nahant to northern cape cod bay, massachusetts, 2013.

\bibitem{Pendleton2019-ci}
Elizabeth~A Pendleton, Edward~M Sweeney, and Laura~L Brothers.
\newblock Optimizing an inner-continental shelf geologic framework investigation through data repurposing and machine learning.
\newblock {\em Geosciences (Basel)}, 9(5):231, May 2019.

\bibitem{Pendleton2015Nahant}
Elizabeth~E Pendleton, Walter~A Barnhardt, Wayne~E Baldwin, David~S Foster, William~C Schwab, Brian~D Andrews, and Seth~D Ackerman.
\newblock Sea-floor texture and physiographic zones of the inner continental shelf from salisbury to nahant, massachusetts, including the merrimack embayment and western massachusetts bay, 2015.

\bibitem{Pendleton2018Nantucket}
Elizabeth~P Pendleton, Wayne~E Baldwin, David~S Foster, Seth Ackerman, Brian~D Andrews, and Laura Brothers.
\newblock Geospatial data layers of shallow geology, sea-floor texture, and physiographic zones from the inner continental shelf of martha's vineyard from aquinnah to wasque point, and nantucket from eel point to great point, 2018.

\bibitem{peng2022out}
Xi~Peng, Fengchun Qiao, and Long Zhao.
\newblock Out-of-domain generalization from a single source: An uncertainty quantification approach.
\newblock {\em IEEE Transactions on Pattern Analysis and Machine Intelligence}, 46(3):1775--1787, 2022.

\bibitem{qiao2023topologyaware}
Fengchun Qiao and Xi~Peng.
\newblock Topology-aware robust optimization for out-of-distribution generalization.
\newblock In {\em The Eleventh International Conference on Learning Representations}, 2023.

\bibitem{qiao2020learning}
Fengchun Qiao, Long Zhao, and Xi~Peng.
\newblock Learning to learn single domain generalization.
\newblock In {\em Proceedings of the IEEE/CVF conference on computer vision and pattern recognition}, pages 12556--12565, 2020.

\bibitem{qin2021ieee}
Xiaoming Qin, Xiaowen Luo, Ziyin Wu, and Jihong Shang.
\newblock Optimizing the sediment classification of small side-scan sonar images based on deep learning.
\newblock {\em IEEE Access}, 9:29416--29428, 2021.

\bibitem{radford2021learning}
Alec Radford, Jong~Wook Kim, Chris Hallacy, Aditya Ramesh, Gabriel Goh, Sandhini Agarwal, Girish Sastry, Amanda Askell, Pamela Mishkin, Jack Clark, Gretchen Krueger, and Ilya Sutskever.
\newblock Learning transferable visual models from natural language supervision, 2021.

\bibitem{Raineault2012-py}
Nicole~A Raineault, Arthur~C Trembanis, and Douglas~C Miller.
\newblock Mapping benthic habitats in delaware bay and the coastal atlantic: Acoustic techniques provide greater coverage and high resolution in complex, {Shallow-Water} environments.
\newblock {\em Estuaries Coast.}, 35(2):682--699, March 2012.

\bibitem{ren2023pixellm}
Zhongwei Ren, Zhicheng Huang, Yunchao Wei, Yao Zhao, Dongmei Fu, Jiashi Feng, and Xiaojie Jin.
\newblock Pixellm: Pixel reasoning with large multimodal model, 2023.

\bibitem{ronneberger2015u}
Olaf Ronneberger, Philipp Fischer, and Thomas Brox.
\newblock U-net: Convolutional networks for biomedical image segmentation.
\newblock In {\em Medical image computing and computer-assisted intervention--MICCAI 2015: 18th international conference, Munich, Germany, October 5-9, 2015, proceedings, part III 18}, pages 234--241. Springer, 2015.

\bibitem{schuhmann2022laion5b}
Christoph Schuhmann, Romain Beaumont, Richard Vencu, Cade Gordon, Ross Wightman, Mehdi Cherti, Theo Coombes, Aarush Katta, Clayton Mullis, Mitchell Wortsman, Patrick Schramowski, Srivatsa Kundurthy, Katherine Crowson, Ludwig Schmidt, Robert Kaczmarczyk, and Jenia Jitsev.
\newblock Laion-5b: An open large-scale dataset for training next generation image-text models, 2022.

\bibitem{schuhmann2021laion400m}
Christoph Schuhmann, Richard Vencu, Romain Beaumont, Robert Kaczmarczyk, Clayton Mullis, Aarush Katta, Theo Coombes, Jenia Jitsev, and Aran Komatsuzaki.
\newblock Laion-400m: Open dataset of clip-filtered 400 million image-text pairs, 2021.

\bibitem{Sethuraman2024MachineLF}
Advaith~Venkatramanan Sethuraman, Anja Sheppard, Onur Bagoren, Christopher Pinnow, Jamey Anderson, T.~Havens, and Katherine~A. Skinner.
\newblock Machine learning for shipwreck segmentation from side scan sonar imagery: Dataset and benchmark.
\newblock {\em ArXiv}, abs/2401.14546, 2024.

\bibitem{sharma-etal-2018-conceptual}
Piyush Sharma, Nan Ding, Sebastian Goodman, and Radu Soricut.
\newblock Conceptual captions: A cleaned, hypernymed, image alt-text dataset for automatic image captioning.
\newblock In Iryna Gurevych and Yusuke Miyao, editors, {\em Proceedings of the 56th Annual Meeting of the Association for Computational Linguistics (Volume 1: Long Papers)}, pages 2556--2565, Melbourne, Australia, July 2018. Association for Computational Linguistics.

\bibitem{Singh2021TheMD}
Deepak Singh and Matias Valdenegro-Toro.
\newblock The marine debris dataset for forward-looking sonar semantic segmentation.
\newblock {\em 2021 IEEE/CVF International Conference on Computer Vision Workshops (ICCVW)}, pages 3734--3742, 2021.

\bibitem{tan2024largelanguagemodelsdata}
Zhen Tan, Dawei Li, Song Wang, Alimohammad Beigi, Bohan Jiang, Amrita Bhattacharjee, Mansooreh Karami, Jundong Li, Lu~Cheng, and Huan Liu.
\newblock Large language models for data annotation: A survey, 2024.

\bibitem{thawkar2023xraygpt}
Omkar Thawkar, Abdelrahman Shaker, Sahal~Shaji Mullappilly, Hisham Cholakkal, Rao~Muhammad Anwer, Salman Khan, Jorma Laaksonen, and Fahad~Shahbaz Khan.
\newblock Xraygpt: Chest radiographs summarization using medical vision-language models, 2023.

\bibitem{Trembanis2019-gs}
Arthur Trembanis, Alimjan Abla, Ken Haulsee, and Carter DuVal.
\newblock Benthic habitat morphodynamics-using remote sensing to quantify storm-induced changes in nearshore bathymetry and surface sediment texture at assateague national seashore.
\newblock {\em J. Mar. Sci. Eng.}, 7(10):371, October 2019.

\bibitem{Trembanis2013Detailed}
Arthur Trembanis, Carter DuVal, Jonathan Beaudoin, Val Schmidt, Doug Miller, and Larry Mayer.
\newblock A detailed seabed signature from hurricane sandy revealed in bedforms and scour.
\newblock {\em Geochemistry, Geophysics, Geosystems}, 14(10):4334--4340, 2013.

\bibitem{Trembanis2020-cl}
Arthur Trembanis, Mark Lundine, and Kaitlyn McPherran.
\newblock Coastal mapping and monitoring.
\newblock In {\em Reference Module in Earth Systems and Environmental Sciences}. Elsevier, 2020.

\bibitem{Walia2023OptimizedCD}
Jaskaran~Singh Walia and Karthik Seemakurthy.
\newblock Optimized custom dataset for efficient detection of underwater trash.
\newblock {\em ArXiv}, abs/2305.16460, 2023.

\bibitem{wang2019underwater}
Xingmei Wang, Jia Jiao, Jingwei Yin, Wensheng Zhao, Xiao Han, and Boxuan Sun.
\newblock Underwater sonar image classification using adaptive weights convolutional neural network.
\newblock {\em Applied Acoustics}, 146:145--154, 2019.

\bibitem{warakagoda2018transfer}
N~Warakagoda and {\O}ivind Midtgaard.
\newblock Transfer-learning with deep neural networks for mine recognition in sonar images.
\newblock {\em SAS/SAR}, 40:115--122, 2018.

\bibitem{williams2016underwater}
David~P. Williams.
\newblock Underwater target classification in synthetic aperture sonar imagery using deep convolutional neural networks.
\newblock In {\em 2016 23rd International Conference on Pattern Recognition (ICPR)}, pages 2497--2502, 2016.

\bibitem{williams2019transfer}
David~P. Williams.
\newblock Transfer learning with sas-image convolutional neural networks for improved underwater target classification.
\newblock In {\em IGARSS 2019 - 2019 IEEE International Geoscience and Remote Sensing Symposium}, pages 78--81, 2019.

\bibitem{Xie2022ADW}
Kaibing Xie, Jian Yang, and Kang Qiu.
\newblock A dataset with multibeam forward-looking sonar for underwater object detection.
\newblock {\em Scientific Data}, 9, 2022.

\bibitem{xu2020underwater}
Yichao Xu, Xingmei Wang, Kunhua Wang, Jiahao Shi, and Wei Sun.
\newblock Underwater sonar image classification using generative adversarial network and convolutional neural network.
\newblock {\em IET Image Processing}, 14(12):2819--2825, 2020.

\bibitem{yang2024lisa}
Senqiao Yang, Tianyuan Qu, Xin Lai, Zhuotao Tian, Bohao Peng, Shu Liu, and Jiaya Jia.
\newblock Lisa++: An improved baseline for reasoning segmentation with large language model, 2024.

\bibitem{ye2018ieee}
Xiufen Ye, Chuanlong Li, Siyuan Zhang, Peng Yang, and Xiang Li.
\newblock Research on side-scan sonar image target classification method based on transfer learning.
\newblock In {\em OCEANS 2018 MTS/IEEE Charleston}, pages 1--6, 2018.

\bibitem{ZevenbergenThorne}
Lyle~W. Zevenbergen and Colin~R. Thorne.
\newblock Quantitative analysis of land surface topography.
\newblock {\em Earth Surface Processes and Landforms}, 12(1):47--56, 1987.

\bibitem{zhang2023gpt4roi}
Shilong Zhang, Peize Sun, Shoufa Chen, Min Xiao, Wenqi Shao, Wenwei Zhang, Yu~Liu, Kai Chen, and Ping Luo.
\newblock Gpt4roi: Instruction tuning large language model on region-of-interest, 2023.

\bibitem{zhang2023pmcvqa}
Xiaoman Zhang, Chaoyi Wu, Ziheng Zhao, Weixiong Lin, Ya~Zhang, Yanfeng Wang, and Weidi Xie.
\newblock Pmc-vqa: Visual instruction tuning for medical visual question answering, 2023.

\bibitem{zhang2024groundhog}
Yichi Zhang, Ziqiao Ma, Xiaofeng Gao, Suhaila Shakiah, Qiaozi Gao, and Joyce Chai.
\newblock Groundhog: Grounding large language models to holistic segmentation, 2024.

\bibitem{zheng2023minigpt5}
Kaizhi Zheng, Xuehai He, and Xin~Eric Wang.
\newblock Minigpt-5: Interleaved vision-and-language generation via generative vokens, 2023.

\bibitem{zhu2023minigpt4}
Deyao Zhu, Jun Chen, Xiaoqian Shen, Xiang Li, and Mohamed Elhoseiny.
\newblock Minigpt-4: Enhancing vision-language understanding with advanced large language models, 2023.

\bibitem{zhu2018underwater}
Keqing Zhu, Jie Tian, and Haining Huang.
\newblock Underwater object images classification based on convolutional neural network.
\newblock In {\em 2018 IEEE 3rd International Conference on Signal and Image Processing (ICSIP)}, pages 301--305, 2018.

\end{thebibliography}
}
\newpage



\section*{Checklist}


\begin{enumerate}

\item For all authors...
\begin{enumerate}
  \item Do the main claims made in the abstract and introduction accurately reflect the paper's contributions and scope?
    \answerYes{}
  \item Did you describe the limitations of your work?
    \answerYes{}
  \item Did you discuss any potential negative societal impacts of your work?
    \answerNo{The primary use of the sonar image dataset is for scientific and environmental monitoring purposes, which inherently aim to support rather than harm societal interests.}
  \item Have you read the ethics review guidelines and ensured that your paper conforms to them?
    \answerYes{}
\end{enumerate}

\item If you are including theoretical results...
\begin{enumerate}
  \item Did you state the full set of assumptions of all theoretical results?
    \answerNA{}
	\item Did you include complete proofs of all theoretical results?
    \answerNA{}
\end{enumerate}

\item If you ran experiments (e.g. for benchmarks)...
\begin{enumerate}
  \item Did you include the code, data, and instructions needed to reproduce the main experimental results (either in the supplemental material or as a URL)?
    \answerYes{}
  \item Did you specify all the training details (e.g., data splits, hyperparameters, how they were chosen)?
    \answerYes{}
	\item Did you report error bars (e.g., with respect to the random seed after running experiments multiple times)?
    \answerYes{}
	\item Did you include the total amount of compute and the type of resources used (e.g., type of GPUs, internal cluster, or cloud provider)?
    \answerYes{}
\end{enumerate}

\item If you are using existing assets (e.g., code, data, models) or curating/releasing new assets...
\begin{enumerate}
  \item If your work uses existing assets, did you cite the creators?
    \answerYes{}
  \item Did you mention the license of the assets?
    \answerYes{}
  \item Did you include any new assets either in the supplemental material or as a URL?
    \answerYes{}
  \item Did you discuss whether and how consent was obtained from people whose data you're using/curating?
    \answerNA{}
  \item Did you discuss whether the data you are using/curating contains personally identifiable information or offensive content?
    \answerNA{}
\end{enumerate}

\item If you used crowdsourcing or conducted research with human subjects...
\begin{enumerate}
  \item Did you include the full text of instructions given to participants and screenshots, if applicable?
    \answerNA{}
  \item Did you describe any potential participant risks, with links to Institutional Review Board (IRB) approvals, if applicable?
    \answerNA{}
  \item Did you include the estimated hourly wage paid to participants and the total amount spent on participant compensation?
    \answerNA{}
\end{enumerate}

\end{enumerate}


\end{document}


\maketitle
\appendix


\section{Datasheets for Datasets}
\label{S3}
\subsection{Motivation}
\label{S3_1}
\textbf{For what purpose was the dataset created?}
The dataset was created to further advance machine learning techniques in the field of marine science. 

\textbf{Who created the dataset (e.g., which team, research group) and on behalf of which entity (e.g., company, institution, organization)?}
The dataset was created by the Deep-REAL and CSHEL labs at the University of Delaware. The sources of the data are from USGS and NOAA.

\textbf{Who funded the creation of the dataset?}
The Department of Defense funded the project under the DEPSCoR Award.

\subsection{Composition}
\label{S3_2}
\textbf{What do the instances that comprise the dataset represent (e.g., documents, photos, people, countries)?}
An instance is a sonar image (2D grid data), containing different geographic layers, each of which is a channel of the image.

\textbf{How many instances are there in total (of each type, if appropriate)?}
Our dataset contains 696K sonar images, 827K segmentation masks, 696K general language descriptions and 7M question-answer pairs.

\textbf{Does the dataset contain all possible instances or is it a sample (not necessarily random) of instances from a larger set?}
It contains all possible instances from the raw collected data.

\textbf{What data does each instance consist of?}
For \texttt{SeafloorAI}, each instance, at most, consists of backscatter signal, bathymetry, slope, rugosity, geo-location, ground-truth annotations such as sediment, physiographic zone, habitat, fault and fold. For \texttt{SeafloorGenAI}, each instance also contains some question-answer pairs and an associated language description.

\textbf{Is there a label or target associated with each instance?}
Not quite, there are image patches without annotations of some or all mapping layers. This is inherently due to the collected raw data.

\textbf{Is any information missing from individual instances?}
No.

\textbf{Are relationships between individual instances made explicit (e.g., users’ movie ratings, social network links)?}
Not applicable.

\textbf{Are there recommended data splits (e.g., training, development/validation, testing)?}
Yes.

\textbf{Are there any errors, sources of noise, or redundancies in the dataset?}
There might be noise from the sensors when the data was collected.

\textbf{Is the dataset self-contained, or does it link to or otherwise rely on external resources (e.g., websites, tweets, other datasets)?}
The raw data surveys are available on different sources. We curated and unified them into a single dataset.

\textbf{Does the dataset contain data that might be considered confidential (e.g., data that is protected by legal privilege or by doctor-patient confidentiality, data that includes the content of individuals’ non-public communications)?}
No.

\textbf{Does the dataset contain data that, if viewed directly, might be offensive, insulting, threatening, or might otherwise cause anxiety?}
No.

\textbf{Does the dataset identify any subpopulations (e.g., by age, gender)?}
No.

\textbf{Is it possible to identify individuals (i.e., one or more natural persons), either directly or indirectly (i.e., in combination with other data) from the dataset?}
No.

\textbf{Does the dataset contain data that might be considered sensitive in any way (e.g., data that reveals race or ethnic origins, sexual orientations, religious beliefs, political opinions or union memberships, or locations; financial or health data; biometric or genetic data; forms of government identification, such as social security numbers; criminal history)?}
No.

\subsection{Collection Process}
\label{S3_3}
\textbf{How was the data associated with each instance acquired? Was the data directly observable (e.g., raw text, movie ratings), reported by subjects (e.g., survey responses), or indirectly inferred/derived from other data (e.g., part-of-speech tags, model-based guesses for age or language)?}
The data was observable; these were raw data collected by the USGS and NOAA, which were then processed by the authors.

\textbf{What mechanisms or procedures were used to collect the data (e.g., hardware apparatuses or sensors, manual human curation, software programs, software APIs)?}
The raw data was collected by sonar sensors. We downloaded the available data from the websites through the browser.

\textbf{If the dataset is a sample from a larger set, what was the sampling strategy (e.g., deterministic, probabilistic with specific sampling probabilities)?}
Not applicable.

\textbf{Who was involved in the data collection process (e.g., students, crowdworkers, contractors) and how were they compensated (e.g., how much were crowdworkers paid)?}
Students were involved in the data collection process, \textit{i.e.}, downloading data from the public data repositories.

\textbf{Over what timeframe was the data collected?}
The raw data was collected from 2004 to 2024.

\textbf{Were any ethical review processes conducted (e.g., by an institutional review board)?}
No.

\textbf{Did you collect the data from the individuals in question directly, or obtain it via third parties or other sources (e.g., websites)?}
The data was obtained from public USGS and NOAA websites. The source of each data survey is included in our references.

\textbf{Were the individuals in question notified about the data collection?}
Not applicable. The data was publicly available.

\textbf{Did the individuals in question consent to the collection and use of their data?}
Not applicable. The data was publicly available.

\textbf{If consent was obtained, were the consenting individuals provided with a mechanism to revoke their consent in the future or for certain uses?}
Not applicable. The data was publicly available.

\textbf{Has an analysis of the potential impact of the dataset and its use on data subjects (e.g., a data protection impact analysis) been conducted?}
Not applicable.

\subsection{Preprocessing/cleaning/labeling}
\label{S3_4}
\textbf{Was any preprocessing/cleaning/labeling of the data done (e.g., discretization or bucketing, tokenization, part-of-speech tagging, SIFT feature extraction, removal of instances, processing of missing values)?}
Reprojection, rasterization, noise removal or smoothing, patchifying, and interpolation of missing data are the major data preprocessing steps involved.

\textbf{Was the “raw” data saved in addition to the preprocessed/cleaned/labeled data (e.g., to support unanticipated future uses)?}
The raw data is available online, with URLs included in the references.

\textbf{Is the software that was used to preprocess/clean/label the data available?}
Yes, we also make our processing code available so that people could contribute to the data pool.

\subsection{Uses}
\label{S3_5}
\textbf{Has the dataset been used for any tasks already?}
No.

\textbf{Is there a repository that links to any or all papers or systems that use the dataset?}
No.

\textbf{What (other) tasks could the dataset be used for?}
Currently, the data can only be used for semantic segmentation, but the unlabeled data can be used to pre-trained a model via self-supervised learning.

\textbf{Is there anything about the composition of the dataset or the way it was collected and preprocessed/cleaned/labeled that might impact future uses?}
No.

\textbf{Are there tasks for which the dataset should not be used?}
No.

\textbf{Any other comments?}
No.

\subsection{Distribution}
\label{S3_6}
\textbf{Will the dataset be distributed to third parties outside of the entity (e.g., company, institution, organization) on behalf of which the dataset was created?}
The dataset is going to be publicly available to all.

\textbf{How will the dataset will be distributed (e.g., tarball on website, API, GitHub)?}
GitHub is going to be the main directory for the dataset, containing documentation, code, and URLs to where the data is stored.

\textbf{When will the dataset be distributed?}
Before Dec 9th, the date of the conference.

\textbf{Will the dataset be distributed under a copyright or other intellectual property (IP) license, and/or under applicable terms of use (ToU)?}
It will be distributed under the CC-BY 4.0 license.

\textbf{Have any third parties imposed IP-based or other restrictions on the data associated with the instances?}
No.

\textbf{Do any export controls or other regulatory restrictions apply to the dataset or to individual instances?}
No.

\subsection{Maintenance}
\label{S3_7}
\textbf{Who will be supporting/hosting/maintaining the dataset?}
The dataset will be hosted on Zenodo. The first author will be maintaining and updating the dataset.

\textbf{How can the owner/curator/manager of the dataset be contacted (e.g., email address)?}
The contact information is included in the author list.

\textbf{Is there an erratum?}
No.

\textbf{Will the dataset be updated (e.g., to correct labeling errors, add new instances, delete instances)?}
The dataset will continue to be expanded in the future.

\textbf{If the dataset relates to people, are there applicable limits on the retention of the data associated with the instances (e.g., were the individuals in question told that their data would be retained for a fixed period of time and then deleted)?}
Not applicable.

\textbf{Will older versions of the dataset continue to be supported/hosted/maintained?}
Yes.

\textbf{If others want to extend/augment/build on/contribute to the dataset, is there a mechanism for them to do so?}
Yes, we will make our code publicly available so other people can contribute.


\section{Appendix}

\begin{table}[htb!]
\centering
\resizebox{\linewidth}{!}{
    \begin{tabular}{p{0.25\linewidth} | p{0.95\linewidth}}
    \toprule
         \textbf{Geological Feature} & \textbf{Definition} \\
    \midrule
         Backscatter & The measure of the intensity of sound waves or radar signals reflected back from the seabed or other underwater objects. Used to infer bottom type and other properties. \\
    \midrule
         Bathymetry & The measurement of the depth of water bodies; essentially the underwater equivalent of topography, mapping the shape and depth of the ocean floor. \\
    \midrule
         Slope & The steepness or incline of the seabed, calculated as the change in elevation over a specific distance. Used in predicting sediment transport and erosion. \\
    \midrule
         Rugosity & The measure of the roughness or irregularity of the seafloor's surface texture, often used in ecological studies to assess habitat complexity. \\
    \midrule
         Sediment & Particulate organic and inorganic material that settles at the bottom of a water body. Its composition can vary widely from clay to gravel. \\
    \midrule
        Physiographic Zones & Broad geographic regions characterized by a distinctive landscape and geological features, which are often similar in terms of their formation and development. \\
    \midrule
        Habitat (Abiotic) & The non-living environmental factors that influence ecosystem structures, such as water depth, temperature, and salinity.\\
    \midrule
        Faults & A fracture or zone of fractures between two blocks of rock, which can lead to relative movement and displacement that is often measurable.\\
    \midrule
        Folds & A bend or series of bends in rock strata or other planar surfaces, resulting from compressive forces that cause the material to buckle and deform permanently. \\
    \bottomrule
    \end{tabular}
}
    \vspace{2pt}
    \caption{Definition of geological features that involve in our seafloor mapping dataset.}
    \label{tab:geo-features}
\end{table}

\begin{table}
\centering
\resizebox{\linewidth}{!}{
    \begin{tabular}{p{0.15\linewidth} | p{0.95\linewidth}}
    \toprule
         \textbf{Region Index} & \shortstack[c]{\textbf{Geological Data Surveys}} \\
    \midrule
         Region 1 & Offshore of Arcata, CA~\cite{Cochrane2024-ju}, Eureka, CA~\cite{Cochrane2024-uh}, Eel River, CA~\cite{Cochrane2023-ua}, and Cape Mendocino, CA~\cite{Cochrane2023-cg} \\
    \midrule
        Region 2 & Offshore of Salt Point, CA~\cite{Johnson2015-dw}, Fort Ross, CA~\cite{Johnson2015-ga}, Bodega Head, CA~\cite{Johnson2015-fe}, Tomales Point, CA~\cite{Johnson2015-ir}, Point Reyes, CA~\cite{Watt2015-ui}, Drakes Bay, CA~\cite{Watt2015-dc}, Bolinas, CA~\cite{Cochrane2015-je}, San Francisco, CA~\cite{Cochrane2015-em}, Pacifica, CA~\cite{Edwards2015-gg}, Half Moon Bay, CA~\cite{Cochrane2014-cd}, San Gregorio, CA~\cite{Cochrane2014-ui}, Pigeon Point, CA~\cite{Cochrane2015-gt}, Scott Creek, CA~\cite{Cochrane2015-mx}, Santa Cruz, CA~\cite{Cochrane2015-jp}, and Aptos, CA~\cite{Cochrane2016-zw}\\
    \midrule
        Region 3 & Offshore of Point Buchon, CA~\cite{Cochrane2022-td}, Point Estero, CA~\cite{Cochrane2022-qt}, and Morro Bay, CA~\cite{Cochrane2022-ev} \\
    \midrule
        Region 4 & Offshore of Point Conception, CA~\cite{Johnson2018-kq}, Gaviota, CA~\cite{Johnson2018-jw}, Refugio Beach, CA~\cite{Johnson2015-jn}, Coal Oil Point, CA~\cite{Johnson2014-wv}, Santa Barbara, CA~\cite{Johnson2013-na}, Carpinteria, CA~\cite{Johnson2013Carpinteria}, Ventura, CA~\cite{Johnson2013Ventura}, and Hueneme Canyon, CA~\cite{Johnson2013Hueneme} \\
    \midrule
        Region 5 & Inner Continental Shelf from Nahant to Northern Cape Cod Bay, MA~\cite{Pendleton2013CapeCod}, and from Salisbury to Nahant, MA~\cite{Pendleton2015Nahant}\\
    \midrule
        Region 6 & Buzzards Bay and Vineyard Sound, MA~\cite{Ackerman2015-bk} \\
    \midrule
        Region 7 & Inner Continental Shelf of Martha’s Vineyard from Aquinnah to Wasque Point, and Nantucket from Eel Point to Great Point, MA~\cite{Pendleton2018Nantucket}\\
    \midrule
        Region 8 & NOAA Survey H11554~\cite{H11554}, H11555~\cite{H11555}, H11647~\cite{H11647}, H11648~\cite{H11648}, H11649~\cite{H11649}, H11650~\cite{H11650}, H11872~\cite{H11872}, H11873~\cite{H11873}, H11874~\cite{H11874}, H11992~\cite{H11992}, H12001~\cite{H12001}, H12002~\cite{H12002}, H12003~\cite{H12003}, H12091~\cite{H12091}, H12092~\cite{H12092}, H12093~\cite{H12093}, H12094~\cite{H12094}, H12160~\cite{H12160}, H12161~\cite{H12161}, H12336~\cite{H12336}, H12337~\cite{H12337}, H12338~\cite{H12338}, H12339~\cite{H12339},
        H12394~\cite{H12394}, H12395~\cite{H12395}, H12396~\cite{H12396}, and H12397~\cite{H12397}\\
    \midrule
        Region 9 & Long Bay from Little River to Winyah Bay, SC~\cite{Baldwin2004SC} \\

    \bottomrule
    \end{tabular}
}
    \caption{Region Index and its corresponding hydrographic surveys in the dataset. Our dataset includes 62 surveys across 9 regions on both the U.S coasts.}
    \label{tab:surveys}
\end{table}

\subsection{Glossary for Geological Features}

We provide the definition of the geological features covered in the paper in Tab.~\ref{tab:geo-features}.

\subsection{Details on public data surveys}
Table~\ref{tab:surveys} includes the name of the original 62 USGS/NOAA data surveys for each region. Our data pool is going to be expanded further as more data sources are available for processing.

\textbf{Discussion on geographical diversity.} Despite consisting only 9 regions on the U.S coasts, \texttt{SeafloorAI}, which we see as a starting point towards a goal of more extensive global settings, is still very representative of geologically diverse environments. We have data gathered from both an active (West Coast) and passive (East Coast) plate boundary with varied tectonic history and lithologic composition. Furthermore, within the east coast dataset we have a range spanning from the Middle Atlantic bight with a wide continental shelf all the way up into New England which includes direct glacial impacts on the coastline and seabed. The areas represent similar settings around the world~\cite{inmanOnthe,Davis2020-rx}. For instance the East Coast of the US as an Amero-trailing edge tectonic setting is very similar to that found off the east coast of South America. The West Coast of the US, which is an active plate boundary with a narrow shelf, is also similar to those found along the west coast of South American and other similar plate boundary settings. There are admittedly more geological settings that we have hopes of gathering but we must work first with what is available and in this paper we seek to develop and demonstrate the effectiveness of the framework that could be acquired readily at this time. Our goal is to develop and test this framework and by illustrating the effectiveness and benefits in using this framework to encourage further adoption to expand the representative datasets.

\subsection{Formulas for Bathymetric Derivatives}

\textbf{Zvenbergen \& Throne Slope Formula.} Slope, or \textit{steepness}, of the terrain, can be computed as the gradients of the elevation. For a pixel at $(i,j)$ in a digital elevation model (DEM) or the bathymetry raster file in our case, $Z_{i,j}$ denotes the elevation at that pixel. The slope $\theta$ is calculated using the following steps:

1. Compute the finite differences:

The elevation differences along the $x-$direction $(\partial Z/\partial X)$ and $y-$direction $(\partial Z/\partial Y)$ are computed by averaging the differences between the central pixel and its eight immediate neighbors, equivalent to the locality of a $3\times 3$ sub-matrix:
\begin{equation}
    \frac{\partial Z}{\partial X} = \frac{Z_{i-1,j-1} + 2Z_{i-1,j} + Z_{i-1,j+1} - Z_{i+1,j-1} - 2Z_{i+1,j} - Z_{i+1,j+1}}{8 \times \text{pixel resolution}} \\
\end{equation}
\begin{equation}
    \frac{\partial Z}{\partial Y} = \frac{Z_{i-1,j-1} + 2Z_{i,j-1} + Z_{i+1,j-1} - Z_{i-1,j+1} - 2Z_{i,j+1} - Z_{i+1,j+1}}{8 \times \text{pixel resolution}}
\end{equation}
2. Compute slope:

The slope $\theta$ is then calculated using the formula:
\begin{equation}
    \theta = \arctan \Bigg(\sqrt{\Big(\frac{\partial Z}{\partial X}\Big)^2 + \Big(\frac{\partial Z}{\partial Y}\Big)^2}\Bigg)
\end{equation}
Finally, we convert the result to degrees: $\theta_{\text{deg}} = \theta \times \frac{180}{\pi}$

\textbf{Rugosity Formula.} Rugosity, or \textit{roughness}, of an area can be calculated by taking the ratio of the surface area over the planar area. 
Given a DEM, the terrain is represented by a grid of elevation points, where each point is denoted by $Z$.
The slope at each point affects how much larger the actual surface area of that point is compared to its projection on a horizontal plane.

For a small surface element with sides $dX$ and $dY$ in the horizontal plane, the actual surface area $dA$ of this element on the terrain, which is not flat due to the slope, can be derived from the gradient of the elevation. Here are the steps to calculate the actual surface area:

1. Calculate the gradients:

Similar to the slope formula above, the gradient of the elevation $Z$ at any point is given by the vector $\nabla Z = \Big( \frac{\partial Z}{\partial X}, \frac{\partial Z}{\partial Y} \Big)$.

2. Compute the differential surface element:

The differential surface element $dA$ on a sloped surface can be calculated from the magnitudes of the partial derivatives:
\begin{equation}
    dA = \sqrt{1 + \Big(\frac{\partial Z}{\partial X}\Big)^2 + \Big(\frac{\partial Z}{\partial Y}\Big)^2} dXdY
\end{equation}

3. Integrate over the entire area:

We integrate $dA$ over the area of interest to find the total surface area $A$ of the terrain:
\begin{equation}
    A = \iint \sqrt{1 + \Big(\frac{\partial Z}{\partial X}\Big)^2 + \Big(\frac{\partial Z}{\partial Y}\Big)^2} dXdY
\end{equation}

4. Derive rugosity:

Finally, we take the ratio of the surface area $A$ over the planar area to derive Rugosity:
\begin{equation}
    R = \frac{A}{\text{pixel resolution}^2}
\end{equation}

\subsection{More on Human Evaluation for Language Annotations}
In this section, we provide more details on the iterative human expert evaluation and prompt refinement procedure for language annotations.
We define the three criteria for evaluation as follows:
\begin{enumerate}
    
    \item \textbf{Factual consistency}: the information of geological features contained in the descriptions is consistent with the provided analytical indicators.
    \item \textbf{Factual completeness:} the coverage of geological features in the descriptions with respect to the provided analytical indicators.
    \item \textbf{Coherence:} the logical and fluid integration of information in the text, allowing it to be easily understood by the marine scientist. This metric is qualitative.
\end{enumerate}
We present our results on \textit{factual consistency} and \textit{factual completeness} across four iterations, and explain how feedback from marine scientists on \textit{coherence} informed our prompt refinements for \texttt{GPT-4}. Overall, \texttt{GPT-4} demonstrated high and stable performance in factual consistency, indicating its ability to accurately reproduce the facts provided in the prompt.

However, the model initially struggled with factual completeness due to its tendency to hallucinate and deviate from the main focus of the task instructions, necessitating iterative prompt refinements. In the first iteration, we employed a simple prompt that included only the key analytical indicators and task instructions. The outputs generated by \texttt{GPT-4} lacked coherence and omitted many details from the indicators, resulting in an average factual completeness score of 48\%.

To address this issue, we explicitly added a task rule to our prompt, instructing \texttt{GPT-4} to include all the analytical indicators in the generated descriptions. This modification significantly enhanced performance, with factual completeness scores increasing to 82\%, 85\%, and 96\% in the subsequent iterations.


Regarding coherence, we updated the prompt by leveraging qualitative feedback from the marine scientists, who noted issues such as "the output is not eloquent" and "the output reads more like a list." To address these concerns in subsequent iterations, we incorporated in-context learning (ICL)~\cite{brown2020language}, which provides the LLM with example inputs and outputs for reference.

To construct these examples, we randomly selected 50 samples from the \texttt{SeafloorAI} dataset encompassing various categories. We then asked the scientists to write detailed descriptions based on the input layers (e.g., backscatter, bathymetry) and the annotated segmentation masks of the mapping layers. When generating language annotations for new samples, we included the analytical indicators as the ICL input and the manually written descriptions as the ICL output, as illustrated in Section~\ref{sec:sample-prompt}. This approach helps the LLM mimic domain-specific language, resulting in more coherent and domain-aligned descriptions.

\subsection{Sample Prompt for \texttt{GPT-4}}\label{sec:sample-prompt}
In this section, we present a sample prompt used with \texttt{GPT-4}. Briefly, the prompt's structure includes several in-context learning examples, detailed information about the analytical indicators of the data sample we are annotating, clear task instructions, and specific rules designed to guide the LLM's output according to our desired objectives.


\begin{tcolorbox}[breakable, colback=mycolback, colframe=myframecolor,title=The template of a prompt input to GPT-4 to generate QA pairs]
\begin{lstlisting}
Here are some pairs of example containing the key analytical indicators of a 
sonar image patch and its corresponding domain expert's description:

1. <Input: Analytical Indicators> <Output: Expert's Description>
2. <Input: Analytical Indicators> <Output: Expert's Description>
3. ...

Here are some key analytical indicators extracted from a 224 x 224 sonar 
image. This patch contains 4 input layers (backscatter, bathymetry, slope and 
rugosity) and <# of mapping layers> mapping layers (sediment/physio zone/
habitat/fault/fold):

<Input: Analytical Indicators for new sample>

There are two tasks for you:
(1) Generate a short description describing the geophysical parameters, 
geological composition and spatial distribution of attributes in the image.
(2) Generate 10 question-answer pairs focusing on (sediment/physio zone/
habitat) composition, spatial distribution (based on the provided polygons), 
and general analytical indicators, based on all the information provided.

The generated description and question-answer pairs MUST follow the 
requirements:
- The description MUST refer to what is available in the provided information 
and MUST NOT refer to related concepts outside of the scope.
- The description MUST contain ALL information of the provided geological 
attributes.
- The description MUST NOT be too broad and must be meaningful.
- The answers MUST be concise.
- The output MUST follow the wording style of the samples from the domain 
experts provided at the beginning.
\end{lstlisting}
\end{tcolorbox}

\begin{tcolorbox}[breakable, colback=mycolback, colframe=myframecolor,title=A sample of in-context learning input-output pair for prompting GPT-4]
\begin{lstlisting}
Input:
Depth range: -36.92 meters to -55.00 meters
Backscatter statistics: mean of 119.71, standard deviation of 72.02
Sediment composition: 
- Muddy sand (Sm): 45%
- Gravelly sand (Sg): 9%
- Gravelly rock (Rg): 20%
- Muddy rock (Rm): 26%
Sediment polygons: the polygon format is [x1, y1, x2, y2, ..., xn, yn] for n 
points. The following is the name of the sediment type followed by the 
polygons of each connected component of that sediment in the image.
- Muddy Sand: [223, 50, 79, 74, ..., 223, 50]
- Gravelly Sand: [91, 48, 100, 53, ..., 91, 48], [0, 0, 37, 63, ..., 0, 0]
- Gravelly Rock: [3, 114, 0, 179, ..., 3, 114], [0, 2, 6, 87, ..., 0, 2], ...
- Muddy Rock: [222, 100, 96, 136, ..., 222, 100]
Physiographic Zone polygons:
- Continental/Island Shelf: [223, 0, 123, 43, ..., 223, 0]
- Continental/Island Shore Complex: [0, 116, 1, 186, ..., 0, 116], ...
- Shelf Basin: [189, 7, 110, 18, ..., 189, 7], [0, 11, 51, 68, ..., 0, 11]
----------------------------------------------------------------------------
Output: 
This region shows a heterogenous mix of soft and hardbottom patches. Most of 
the region is dominated by a patch of muddy sand that runs from the lower 
left portion up through the middle and then turns toward the right where it 
narrows between a patch of gravelly rock with higher relief in the upper 
right corner which is part of a shelf basin complex and to the lower right 
where the muddy sand patch is constrained by a continental shore complex 
comprised of muddy rock which is also elevated. The central muddy sand is 
in a deeper depression with water depths ranging between 50-55 m whereas the 
adjacent gravelly rock and muddy rock areas have a shallower depth range of 
between 36-40 m depth. Backscatter within the muddy zones is much lower with 
ranges of around 25-120 compared to the areas with gravelly rock and muddy 
rock which possess much higher backscatter in the range above 150. The slope 
and rugosity is very low in the muddy zones illustrating that these deeper 
areas with fine muddy sediment are flat and seemingly featureless. Whereas, 
the edges of the gravelly rock and muddy rock zones are well delineated by 
areas of high slope and rugosity which represent the shift from a deeper 
flatter zone to the shallower and more rugged terrain in these zones.
\end{lstlisting}
\end{tcolorbox}

\begin{figure}[b!]
    \centering
    \includegraphics[width=\textwidth]{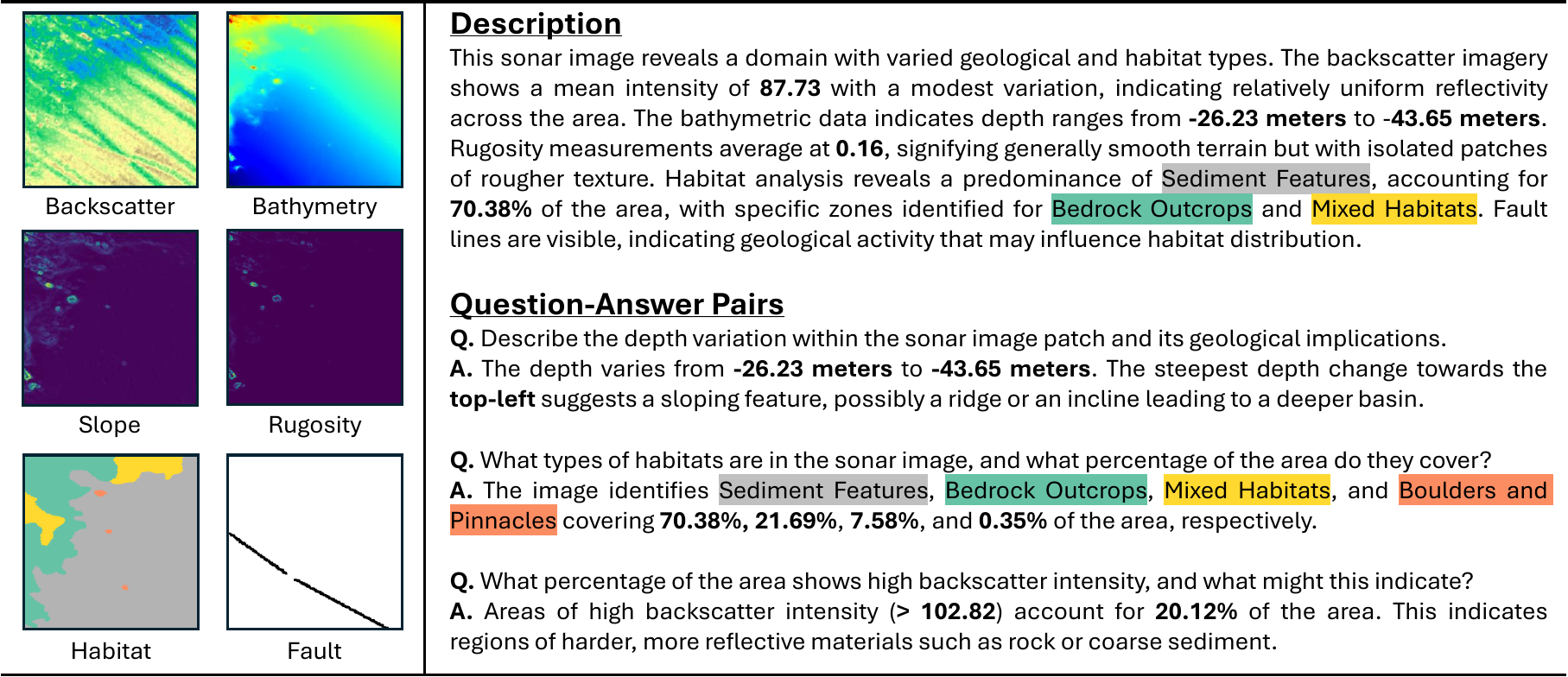}
    \includegraphics[width=\textwidth]{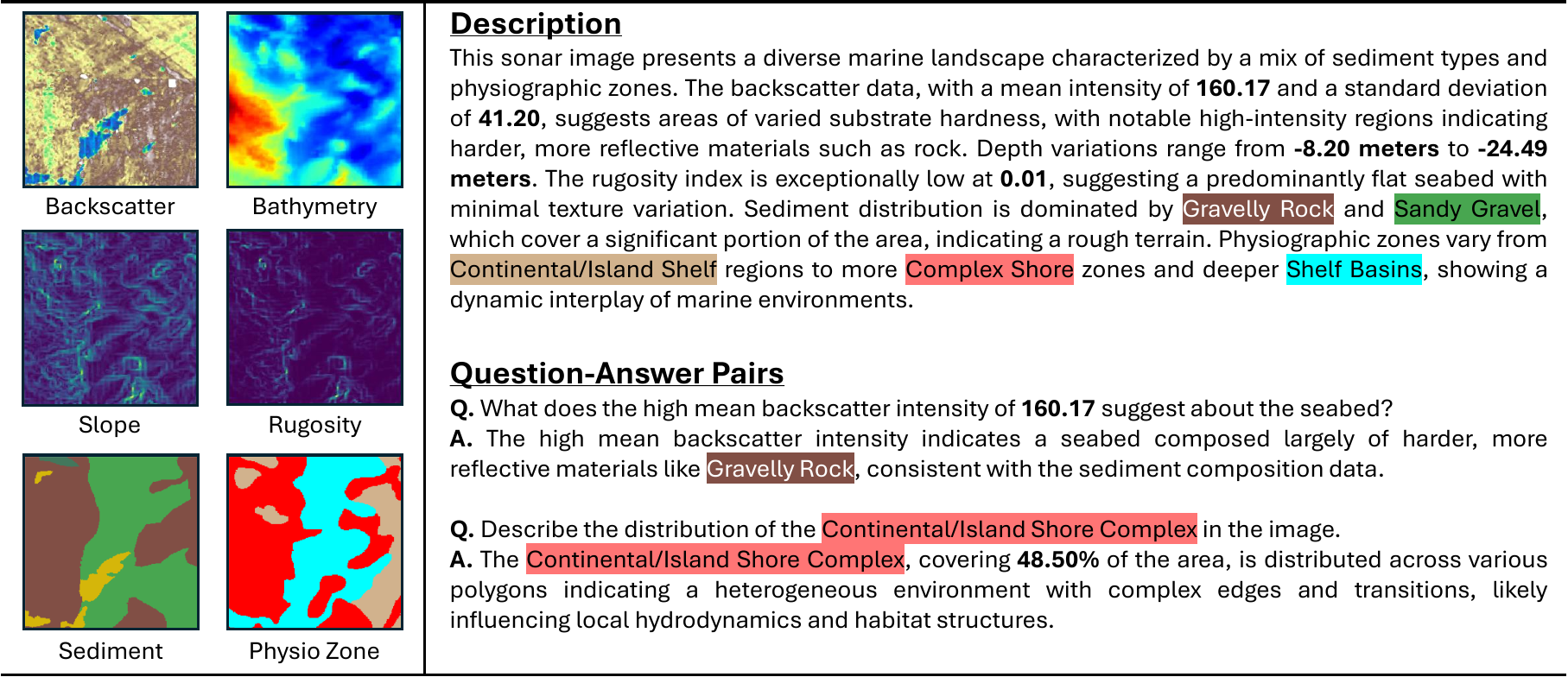}
    \caption{Two additional examples in the \texttt{SeafloorGenAI} dataset, originated from Region 2 (\textbf{top}) and 5 (\textbf{bottom}). It features a \texttt{GPT-4} generated description and question-answer pairs designed to efficiently assist marine scientists in data analysis. The generated description covers all three key analytical indicators.}
    \label{fig:demo}
    \vspace{-5pt}
\end{figure}

\subsection{Additional Examples of the \texttt{SeafloorGenAI} Dataset}

Please refer to Fig.~\ref{fig:demo} for additional examples in the \texttt{SeafloorGenAI} dataset. We include the general descriptions along with some analysis-driven question-answer pairs.

\subsection{More Experiments}
We report some more baseline experiment runs on the binary segmentation tasks on the Fault and Fold layers in this section.

\textbf{Evaluation Metrics.}
We use Dice coefficient and mean IoU for binary segmentation.

\textbf{Data Split.}
Binary segmentation often comes with the problem of class imbalance where the number of background pixels overwhelms that of the object of interest. We choose Region 2 to be the source and Region 4 as target data because Region 2 has more data (Tab.~\ref{tab:splits}). We also employ sample weighting during the training process to alleviate the problem.

\begin{table}[!htb]
\resizebox{0.5\textwidth}{!}{
\begin{tabular}{l|l|l|l}
\toprule
\textbf{Task} & \textbf{Layer} & \multicolumn{1}{c|}{\textbf{Source}}    & \multicolumn{1}{c}{\textbf{Target}} \\
\midrule
\multirow{2}{*}{Binary Segmentation} & Fault & Region 2 & Region 4 \\
& Fold & Region 2 & Region 4 \\
\bottomrule
\end{tabular}}
\caption{Geo-distributed data splits for the binary segmentation tasks on the \texttt{SeafloorAI} dataset.}
\label{tab:splits}
\end{table}

\textbf{Training Details.}
We employ the UNet architecture with different backbones as baselines. 
For binary segmentation, we use more complex backbones, i.e. UNet-ResNet50~\cite{he2015deep} and UNet-DenseNet121~\cite{huang2018densely}, due to the problem of severe class imbalance.

We adopt Focal loss~\cite{lin2018focal} combined with Dice loss for binary segmentation; we set the Focal loss's $\gamma=5.0$. For binary segmentation, we also employ sample weights sampling to alleviate class imbalance.
The weight of each sample with the label of interest (labeled 1) is inversely proportional to the ratio of the number of the foreground pixels to the total number of pixels. The samples with all background pixels are assigned very low weights.
The model was trained using the Adam optimizer~\cite{kingma2017adam}. The learning rate was initially set to 0.001 with a cosine annealing schedule.
We use a batch size of 32 in binary segmentation tasks for 100 epochs, setting the patience to 5 epochs for early stopping.
We perform 3 runs with different random seeds and report the model performance in Tab.~\ref{tab:results2}.
All runs are conducted on a single NVIDIA RTX A6000 GPU.

\textbf{Results.} Tab.~\ref{tab:results2} reports the results on the geo-distributed setting.
Overall, we can see that all baseline models suffer from a signification performance degradation under distribution shift, similar to the multi-class segmentation setting.

\begin{table}[htb!]
\resizebox{0.8\textwidth}{!}{
\begin{tabular}{l|ccc|ccc}
\toprule
\multirow{2}{*}{} & \multicolumn{6}{c}{\textbf{Fault}} \\
\cmidrule{2-7}
& Dice ID & Dice OOD & $\Delta$ Dice & mIoU ID & mIoU OOD & $\Delta$ mIoU \\
\midrule
UNet-DenseNet121 & 90.29 $\pm$ 0.10 & 57.42 $\pm$ 0.92 & -32.87 & 82.32 $\pm$ 0.47 & 52.69 $\pm$ 0.44 & -29.63\\
UNet-ResNet50  & 92.64 $\pm$ 0.54 & 57.18 $\pm$ 0.63 & -35.46 & 86.24 $\pm$ 0.72 & 53.10 $\pm$ 0.23 & -33.14\\

\midrule
\multirow{2}{*}{} & \multicolumn{6}{c}{\textbf{Fold}}  \\
\cmidrule{2-7}
& Dice ID & Dice OOD & $\Delta$ Dice & mIoU ID & mIoU OOD & $\Delta$ mIoU \\
\midrule
UNet-DenseNet121  & 88.84 $\pm$ 0.32 & 54.23 $\pm$ 0.54 & -34.61 & 79.94 $\pm$ 0.56 & 45.89 $\pm$ 0.54 & -34.05\\
UNet-ResNet50  & 92.43 $\pm$ 0.33 & 55.77 $\pm$ 0.67 & -36.66 & 85.93 $\pm$ 0.64 & 47.02 $\pm$ 0.22 & -38.91\\

\bottomrule
\end{tabular}
}
\caption{Performance of the baselines in the geo-distributed setting for binary segmentation.}
\label{tab:results2}
\end{table}

\newpage
{
\small
\bibliographystyle{plain}
\bibliography{Styles/neurips_2024}
}